%% file: author.tex
\documentclass{svproc}

\usepackage{url}

\usepackage{graphicx}
\usepackage{appendix}
\usepackage{caption}
\usepackage{subcaption}
\usepackage{color}
\usepackage{url}

\usepackage[style=nature, backend=biber]{biblatex} 
\newboolean{showcomments}
\setboolean{showcomments}{true}
\ifthenelse{\boolean{showcomments}}
{ \newcommand{\mynote}[3]{
     \fbox{\bfseries\sffamily\scriptsize#1}
       {\small$\blacktriangleright$\textsf{\textcolor{#3}{{\em #2}\bf }}$\blacktriangleleft$}}}
       { \newcommand{\mynote}[2]{}}

\begin{document}
\mainmatter              
\title{Generating Spatial Synthetic Populations Using Wasserstein Generative Adversarial Network: A Case Study with EU-SILC Data for Helsinki and Thessaloniki}
%
\titlerunning{Synthetic Populations by WGAN}  
%
\author{Vanja Falck}
\authorrunning{ } 
%

\institute{Centre for Modelling Social Systems,\\
    NORCE Norwegian Research Center AS,\\
   Universitetsveien 19, Kristiansand, Norway\\
    \email{vafa@norceresearch.no}
}
\maketitle              


\input{0-Abstract/abstract}
\input{1-Introduction/introduction}
\input{2-Methods/methods}
\input{3-Results/results}
\input{4-Discussion/discussion}
\input{5-Conclusion/conclusion}

\section*{Conflicts of Interest} 
The author declare no conflict of interest.

\section*{Acknowledgements}
The work reported here is part of the URBANE project, which has received funding from the European Union’s Horizon Europe Innovation Action under grant agreement No. 101069782.

The author would like to thank Önder Gürcan for his careful review and constructive suggestions, which greatly improved the quality of this paper.

\printbibliography
\input{6-Appendix/appendix}
\end{document}

%% file: 0-Abstract/abstract.tex
\begin{abstract}
Using agent-based social simulations can enhance our understanding of urban planning, public health, and economic forecasting. Realistic synthetic populations with numerous attributes strengthen these simulations. The Wasserstein Generative Adversarial Network, trained on census data like EU-SILC, can create robust synthetic populations. These methods, aided by external statistics or EU-SILC weights, generate spatial synthetic populations for agent-based models. The increased access to high-quality micro-data has sparked interest in synthetic populations, which preserve demographic profiles and analytical strength while ensuring privacy and preventing discrimination. This study uses national data from Finland and Greece for Helsinki and Thessaloniki to explore balanced spatial synthetic population generation. Results show challenges related to balancing data with or without aggregated statistics for the target population and the general under-representation of fringe profiles by deep generative methods. The latter can lead to discrimination in agent-based simulations.

\keywords{synthetic population, deep generative methods, generative adversarial networks, micro-simulations, agent-based models, weighting}
\end{abstract}

%% file: 1-Introduction/introduction.tex
\section{Introduction}



Synthetic populations mimic a real population and are used in micro-simulations \cite{Spatial-Microsimulation-Deterministic-Reweighting-Harland-Clark-Smith-2009} and agent-based models \cite{Synthetic-Population-Varying-Spatial-Scales-Harland-2012}, replicating demographics such as age, gender, and education based on place of residence. However, current methods \cite{Synthetic-Population-Varying-Spatial-Scales-Harland-2012, French-Households-IPF, Microsim-Review-Tanton-2014, Review_Synthetic-Population-Chapuis-2022} face computational challenges as the number of attributes increases \cite{Borysov-Variational-Encoder-Population-2019, WGAN-versus-VAE-zero-cell-Garrido-2020}. A richer attribute profile brings agents closer to fitting complex real-life scenarios \cite{ABM-public-health-complex-systems-Silverman-2021, ABM-complex-wicked-systems-Polhill-2021, ABM-prediction-complex-social-systems-Polhill-2023}. Deep generative methods address this by creating realistic, high-dimensional synthetic populations from original micro-data \cite{Borysov-Variational-Encoder-Population-2019, WGAN-versus-VAE-zero-cell-Garrido-2020}, like EU-SILC \cite{Eurostat-EU-SILC}. Validating these populations presents new challenges due to increasingly complex feature relationships.

EU-SILC provides high-quality census data on living conditions at national and regional levels \cite{Eurostat-EU-SILC}. This dataset is standardised and statistically representative of populations across European countries and collaborative nations. It includes a rich demographic portfolio essential for creating realistic simulation agents. However, access to EU-SILC data is restricted, requiring researchers to apply for permission. While the original data can be used directly in simulations, plain EU-SILC data is biased and only available at a higher regional level \cite{EU-NUTS-Explained}. This bias arises because individual records do not perfectly represent the statistical sample population. A weight is provided to adjust for this misrepresentation. 

Balancing data can be done using the accompanied weight in EU-SILC or applying external demographic profiles fitting the target population \cite{Synthetic-Population-Varying-Spatial-Scales-Harland-2012}. Weights in EU-SILC are floating numbers requiring approximation to integers and a quite heavy upscaling of original data by duplicating original data records. Deep generative methods create data records similar to originals. These replicas are used to match aggregated statistics, avoiding duplicating original records. These balanced datasets next train a deep generative model to produce an utterly synthetic population for use in agent-based models. However, even if the training data is balanced, the models can still generate biased data due to neural networks outputting more frequent examples and fewer examples of fringe profiles. A significant concern is, therefore, the deep generative methods' failure to produce a fair representation of small, vulnerable groups. 

This paper explores strategies for balancing plain high-featured EU-SILC data with either weights or aggregated statistics to produce training data for deep generative networks. The Wasserstein Generative Adversarial Network (WGAN) \cite{GAN-Arjovsky-2017, WGAN-GP-explained-Gulrajani-Arjovsky-2017, WGAN-versus-VAE-zero-cell-Garrido-2020} and EU-SILC cross-sectional data \cite{Eurostat-EU-SILC, METADATA-EU-SILC} are used to create synthetic populations for Helsinki, Finland, and Thessaloniki, Greece. This project uses the same approach as Garrido et al. \cite{WGAN-versus-VAE-zero-cell-Garrido-2020}. However, they did not balance data nor check for the models' ability to represent less favourable groups fairly. This project uses the variable self-perceived health, a strong indicator of quality of life and health, from the EU-SILC data set to profile the extent of misrepresentation.

The synthetic populations are validated using standardised root mean squared error, Pearson's correlation coefficient, R-squared \cite{WGAN-versus-VAE-zero-cell-Garrido-2020} and Bland-Altman plots \cite{Bland-Altman-Method-Critique-Mansournia-2021, French-Households-IPF}.

The contributions of this paper are as follows:

\begin{enumerate}
\item Generation of high-featured spatial synthetic populations using WGAN based on balanced EU-SILC data.
\item Investigation of whether weighting with duplicate originals to train a WGAN model produces high-quality replicas.
\item Examination of whether fitting to external aggregated statistics by imputing original data with WGAN replicas results in high-quality replicas.
\item Consideration of discrimination of fringe groups in the approaches mentioned above.
\item Consideration of validation methods to assess the quality of high-featured synthetic populations.
\end{enumerate}

The paper is structured as follows: Section \ref{sec:Methods} details the methodology and tools for generating synthetic populations, including an overview of the WGAN and its application to replicating and balancing EU-SILC data. Section \ref{sec:Results} presents the results of applying the WGAN approach to Helsinki and Thessaloniki, comparing different balancing strategies. Section \ref{sec:Discussion} discusses the implications, challenges, and benefits of using deep generative methods for synthetic population generation using EU-SILC data, evaluating replication quality and discrimination. Finally, Section \ref{sec:Conclusion} summarises the findings, potential limitations, and suggestions for future research.

%% file: 2-Methods/methods.tex
\section{Methods}
\label{sec:Methods}

Current methods for generating synthetic populations fall into two categories based on data access. The first uses only aggregated statistics to generate random individual data records. The second requires individual survey data and aggregated statistics to match the marginals to correct bias in demographic representations in the survey. These methods are computationally intractable when the number of attributes increases \cite{Synthetic-Population-Varying-Spatial-Scales-Harland-2012, Spatial-Microsimulation-Deterministic-Reweighting-Harland-Clark-Smith-2009, Microsim-Review-Tanton-2014}, and therefore cannot be compared with new generative methods that can offer more realistic and feature-rich synthetic populations. 

EU-SILC provides detailed demographic and standardised living conditions data for all EU countries like Finland and Greece, enabling more complex demographic profiles for realistic synthetic populations. However, data are only available at higher regional levels and not for cities like Helsinki and Thessaloniki. EU-SILC data have a bias accounted for by person weights.

\subsection{EU-SILC Data for Finland and Greece 2022}
EU-SILC data from Finland and Greece for 2022 are obtained from Eurostat \cite{Eurostat-EU-SILC}. 57 variables representing a binary vector of size 294 were included in the training data for both countries. The sample size for Finland was 17515 and for Greece 19480. 

Helsinki provides external aggregated statistics on the combination of age, gender, and education, while Thessaloniki only has separate data for age and gender. Therefore, different methods are needed to balance and adjust the data. First, EU-SILC weights balance the data for Greece. The balancing is done by duplicating the original data of 19480 records according to an approximated integer weight, resulting in a duplicate-imputed dataset of 308559 records. No original data records are removed. Second, external demographic keys are used to balance the demographic profile of Helsinki's population. The fitting to the demographic profile is done by training a WGAN model on the original data for Finland and then from a generated pool of 300000 records, extracting the missing number of persons to fit the profile without removing any original data. The resulting WGAN-imputed dataset is an approximate doubling of the original size of 17515.

The balanced populations are next used as training data for WGAN models to generate synthetic populations for Helsinki and Thessaloniki. The study explores how WGANs can best match the citizen profiles of Helsinki and Thessaloniki, using EU-SILC national micro-data divided into NUTS-1 (Greece) and NUTS-2 (Finland) regions \cite{EU-NUTS-Explained}. While the synthetic population for Greece by default should match the demographic profile of the NUTS-1 region, it may not precisely reflect the demographic profile of Thessaloniki municipality.

Two approaches to balancing are tested:

\begin{itemize}
    \item Approach 1 - fitting external aggregated statistics with WGAN replicas (WGAN-impute)
    \item Approach 2 - using EU-SILC weights by adding copies of original data (duplicate-impute)
\end{itemize}

Weight balancing is the only available approach for Thessaloniki, so validating against the Helsinki population helps assess its usefulness compared to demographic profiling by WGAN for Thessaloniki. See details of data preparations in Appendix \ref{data-preparation}.

\subsection{Wasserstein Generative Adversarial Network}
Generative adversarial networks comprise two neural nets: a critic distinguishing fake from actual data and a generator producing fake data. They compete by the critic assessing data generated by the generator. As the critic improves at discerning real from fake, the generator refines its fakes to deceive. Once trained to convergence, the generator produces convincing "real" data. The Wasserstein generative adversarial network (WGAN) uses Wasserstein distance, RMSEprop for loss, and a gradient penalty function to prevent model collapse. Detailed math of the model is in \cite{GAN-Arjovsky-2017, WGAN-GP-explained-Gulrajani-Arjovsky-2017}, with code akin to the synthetic population's generation in transporting research in \cite{WGAN-versus-VAE-zero-cell-Garrido-2020} and code used in this project in the Appendix \ref{WGAN-Code}.

\subsection{Validation}
Validating synthetic populations is challenging, with no established standards \cite{Review_Synthetic-Population-Chapuis-2022, Microsim-Review-Tanton-2014, Small-Areas-Public-Health-Rahman-Harding-2017, French-Households-IPF}. Validity varies depending on the intended use of the results.

In this paper, the concept of validity comes from quasi-experimental research, categorising it into statistical, construct, internal, and external validity \cite{Quasi-Experimentation-Cook-Campbell}. For synthetic populations, statistical validity refers to maintaining the same statistical properties as the original rather than the strength of statistical conclusions. Construct validity, as the strength of conceptual operationalisation to a particular area of investigation, is beyond this paper's scope. Internal validity concerns issues that the synthetic population induces that otherwise would not be present using the original microdata on a specific problem. In short, internal validity is about the synthetic data being able to take over for original data. Internal validity depends on reliability, which is ensured by any neural network using fixed random seeds. However, issues related to under and over-representing particular profiles in synthetic data threaten internal validity. Shallow statistical properties can be measured by Pearson's correlation coefficient, R-squared and SRMSE, while misrepresentations caused by simultaneous over- and under-representations go under the radar. Neural networks may generate more examples with typical profiles and under-represent fringe cases, threatening internal validity. External validity focuses on how well the synthetic population substitutes for original data in a particular use case.

In this study, the quality of the synthetic population is measured by standardised root mean squared error (SRMSE), Pearson's correlation, and R-squared \cite{WGAN-versus-VAE-zero-cell-Garrido-2020}. Additionally, Bland-Altman plots test and visualise the similarity between original and synthetic data \cite{Bland-Altman-Method-Critique-Mansournia-2021, French-Households-IPF}. Bland-Altman analysis shows how well the synthetic population substitutes the original. Bland-Altman can handle numerous attributes non-linearly, offering a better comparison of complex data structures than the more shallow measures of Pearson's correlation and R-squared. The Bland-Altman analysis plot shows if two methods measure the same phenomenon equivalently \cite{Bland-Altman-Method-Critique-Mansournia-2021}. Outliers suggest that certain variables may differ, requiring further investigation to ensure internal and external validity based on the synthetic population's intended use.

%% file: 3-Results/results.tex
\section{Results}
\label{sec:Results}


The results are organised by presenting the country and region level reproduction measured for Finland and Greece in Figure \ref{fig:compare-synthetic-models-wgan-finland-greece} against the data they are trained on to assess statistical and internal validity.

\begin{figure}
    \centering
\begin{subfigure}{0.32\textwidth}
    \centering
    \includegraphics[width=\textwidth]{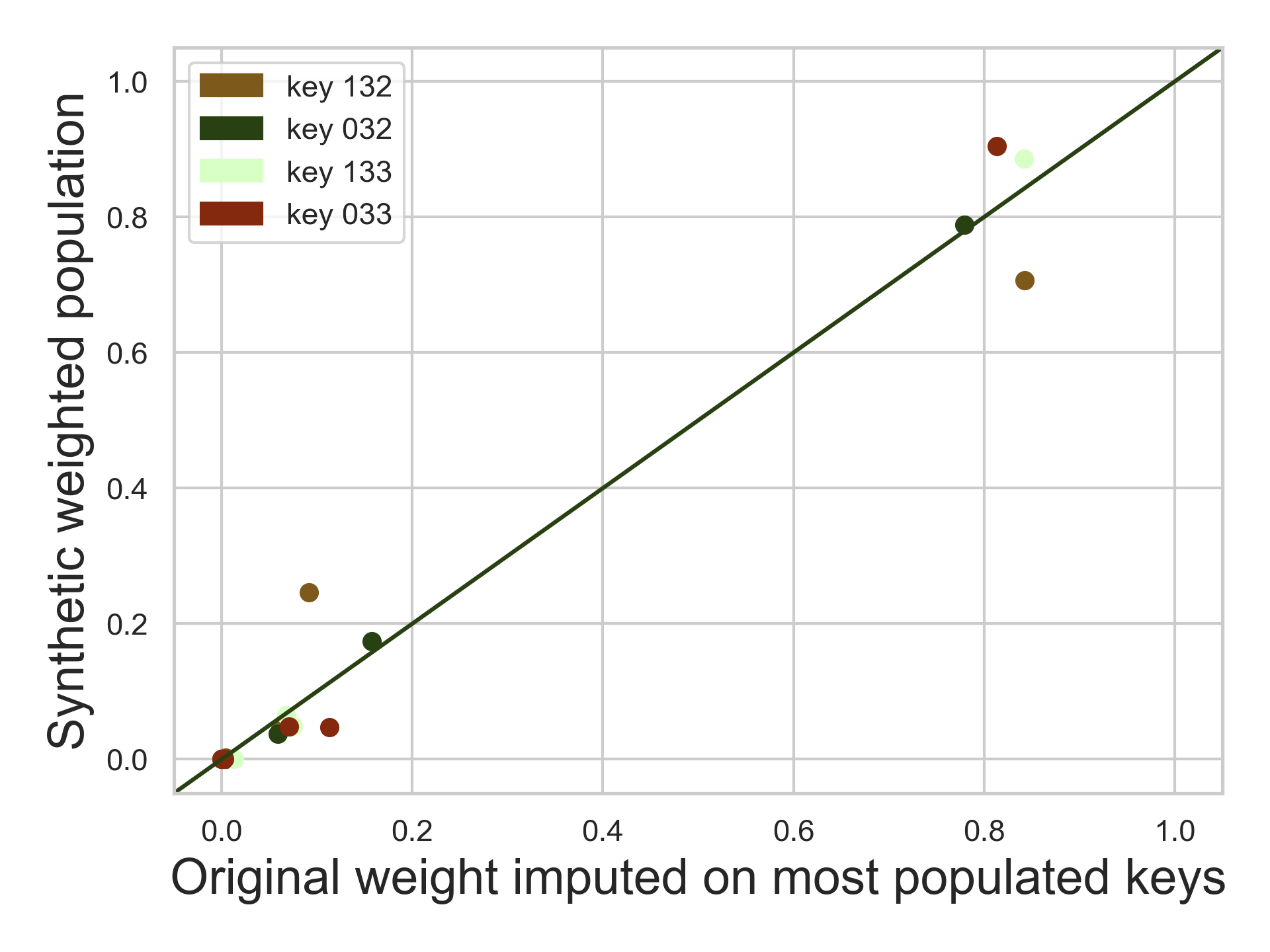}
    \caption[]{Duplicated synthetic to weight-imputed original}
    \label{fig:PH010-weight-to-weight-imputed-original}
    \end{subfigure}
\begin{subfigure}{0.32\textwidth}
    \centering
    \includegraphics[width=\textwidth]{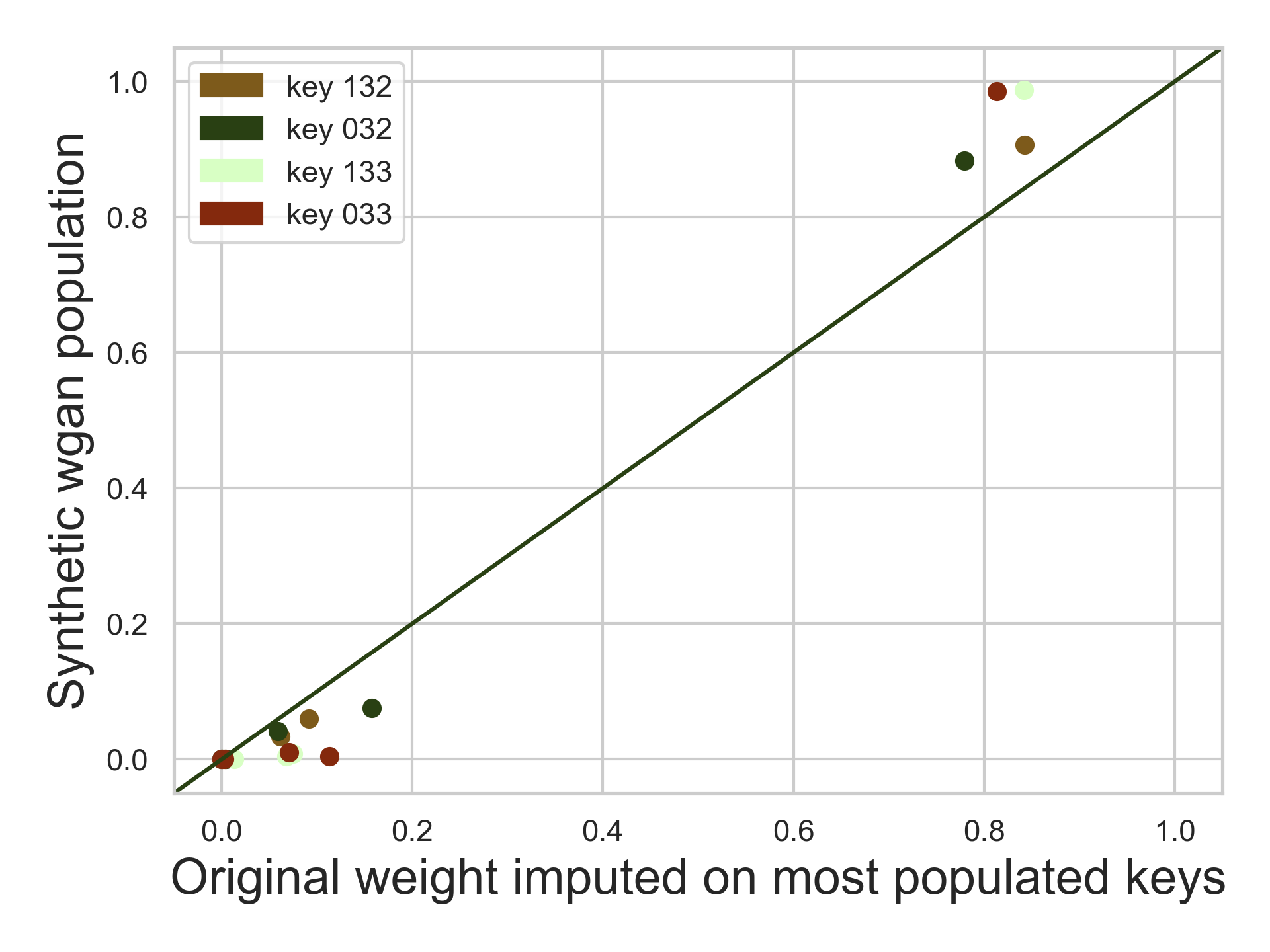}
    \caption[]{WGAN synthetic to weight-imputed original}
    \label{fig:PH010-wgan-to-weight-imputed-original}
    \end{subfigure}
\begin{subfigure}{0.32\textwidth}
    \centering
    \includegraphics[width=\textwidth]{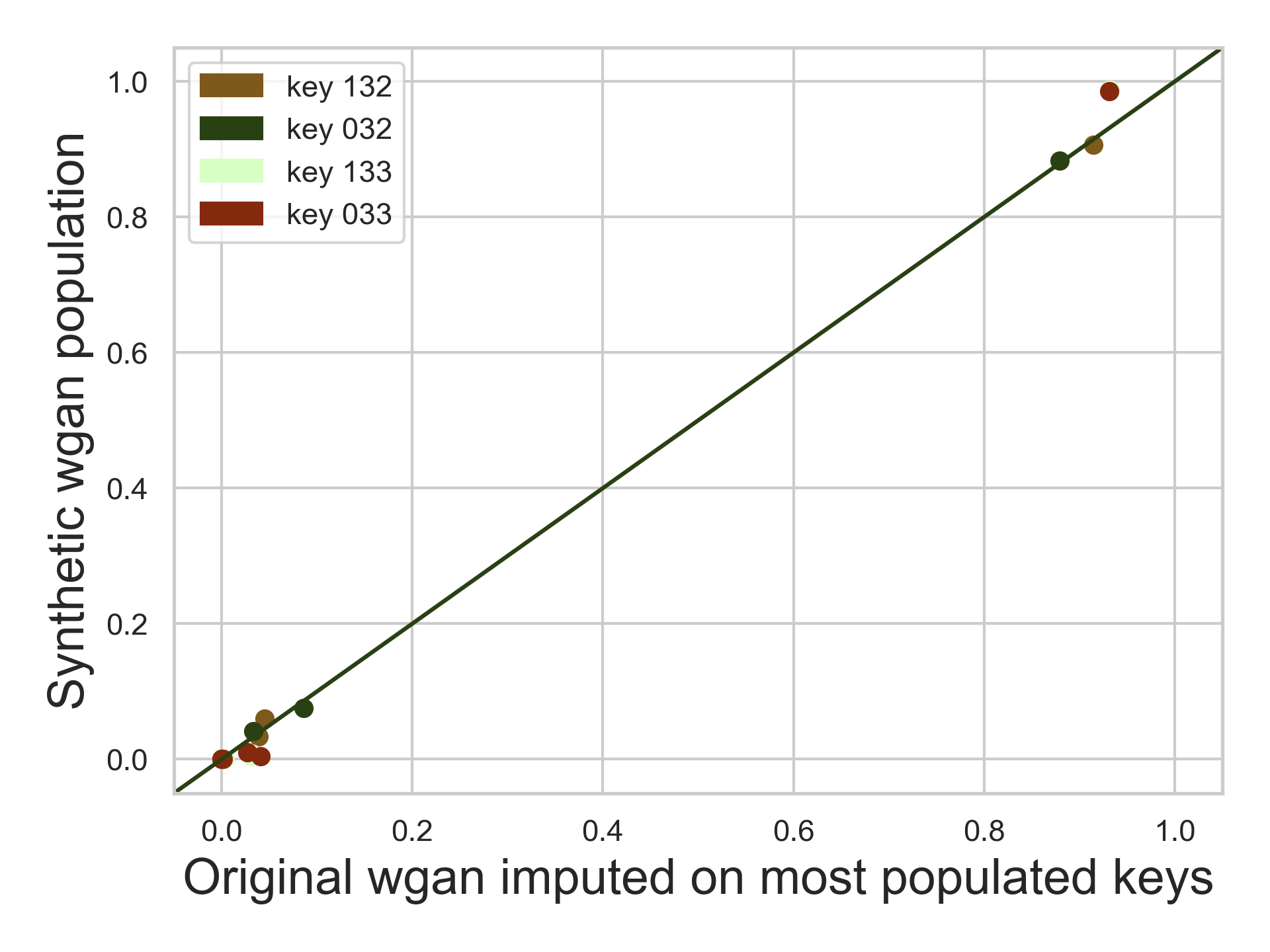}
    \caption[]{WGAN synthetic to wgan-imputed original}
    \label{fig:PH010-wgan-to-wgan-imputed-original}
    \end{subfigure}
\caption[Reproduction of variable self-perceived health (PH010)]{Figure (a) shows self-perceived health (PH010) in synthetic data from training on weight-imputed originals for Finland is compared to weight-imputed original data on the most populated demographic keys. Figure (b) compares synthetic data from training on wgan-imputed originals with weight-imputed original data. Figure (c) compares synthetic data trained on wgan-imputed originals with wgan-imputed original data.}
  \label{fig:PH010-skew-Helsinki}
\end{figure}

Next, the synthetic populations derived from weight-imputed and wgan-imputed for Finland are compared to their training data on the variable self-perceived health (PH010) in Figure \ref{fig:PH010-skew-Helsinki}. A cross-comparison between the synthetic population based on wgan-imputing and the weight-imputed original population is also done on PH010 to assess potential discrimination of the under-representing fringe group in Figure \ref{fig:PH010-wgan-to-weight-imputed-original}.

Finally, the synthetic population derived from weight and wgan-imputed originals for Finland are compared to external aggregated demographic statistics for Helsinki to assess external validity.

\subsection{Helsinki}
Demographic keys of age, gender, and education for Helsinki are obtained from the Finnish Statistical Agency's website. The EU-SILC for Finland is divided into NUTS-2 regions. The Helsinki-Uusimaa region includes Helsinki municipality and is the source for creating the city population. Helsinki and the Uusimaa area are approximately equal in population size, just above 3 million. Helsinki municipality has about 675,000 inhabitants of all ages. The EU-SILC covers people ages 16 and up. This population in 2022 is 568,000.

\begin{figure}
    \centering
\begin{subfigure}{0.22\textwidth}
    \centering
    \includegraphics[width=\textwidth]{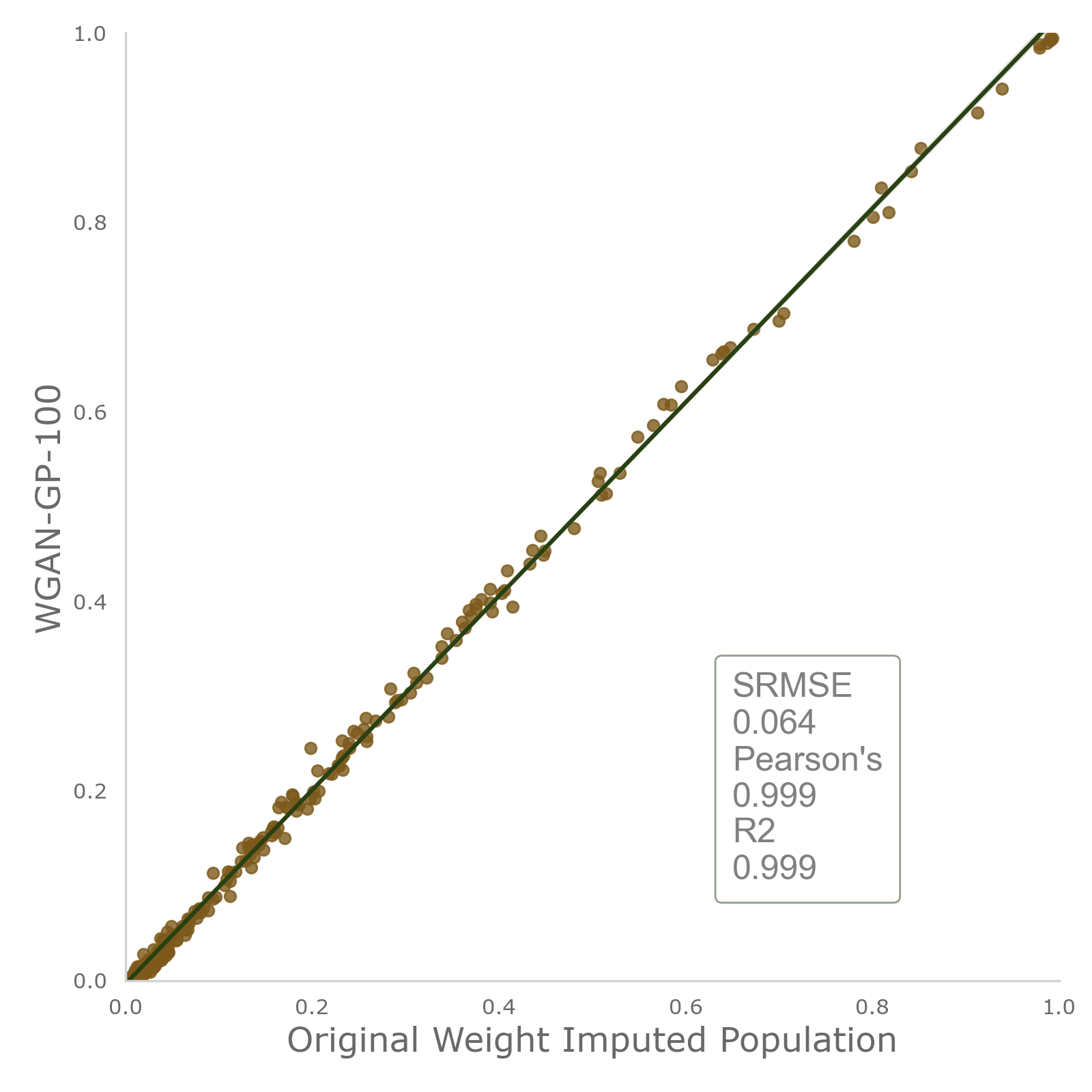}
    \caption[Univariate representation of WGAN on EU-SILC Finland]{Finland}
    \label{fig:finland-full-population}
    \end{subfigure}
\begin{subfigure}{0.22\textwidth}
    \centering
    \includegraphics[width=\textwidth]{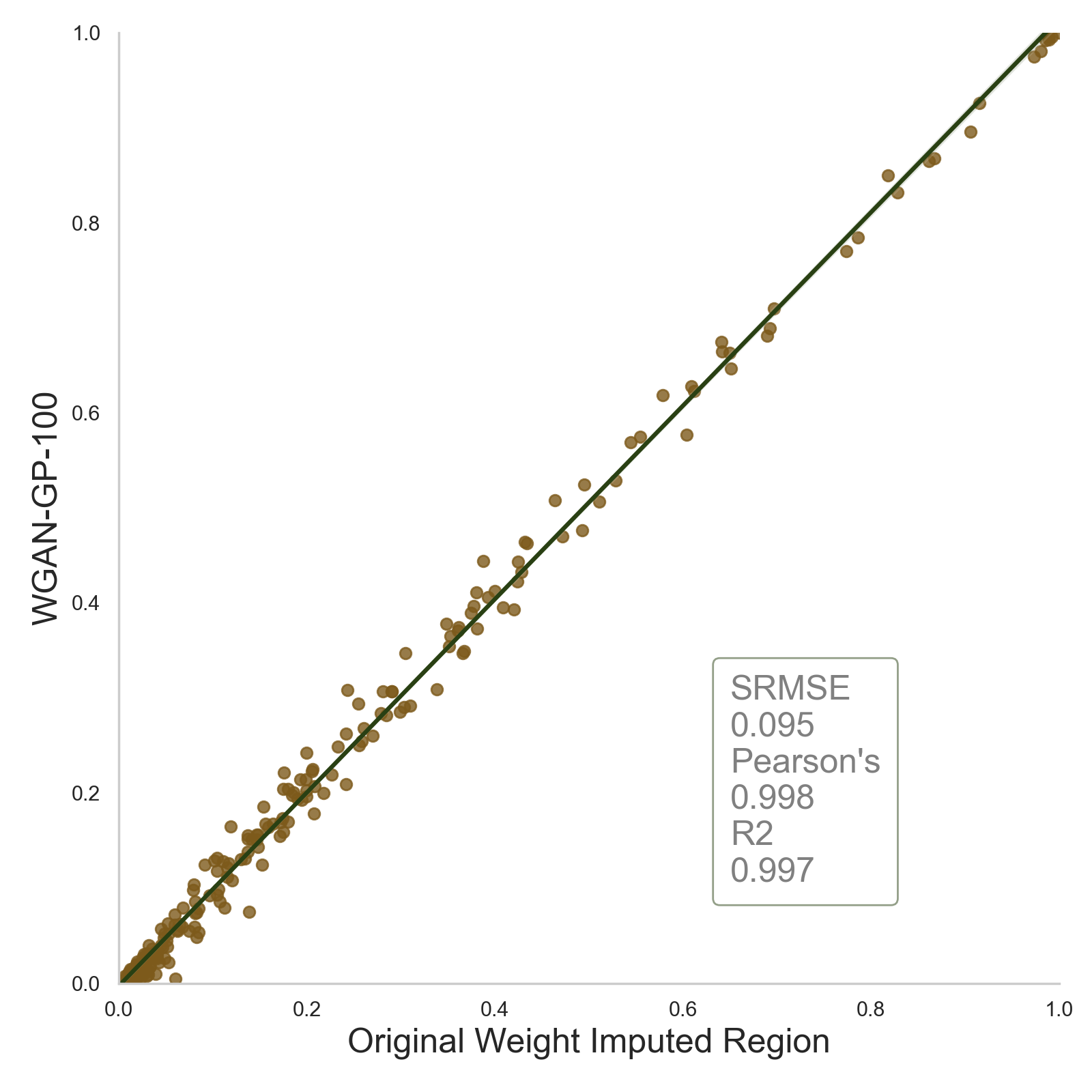}
\caption[Univariate representation of WGAN on EU-SILC Region NUTS-2 including Helsinki from training on complete national level data and extracting region]{Helsinki 1}
    \label{fig:helsinki-weight-imputed-from-population}
    \end{subfigure}
\begin{subfigure}{0.22\textwidth}
    \centering
    \includegraphics[width=\textwidth]{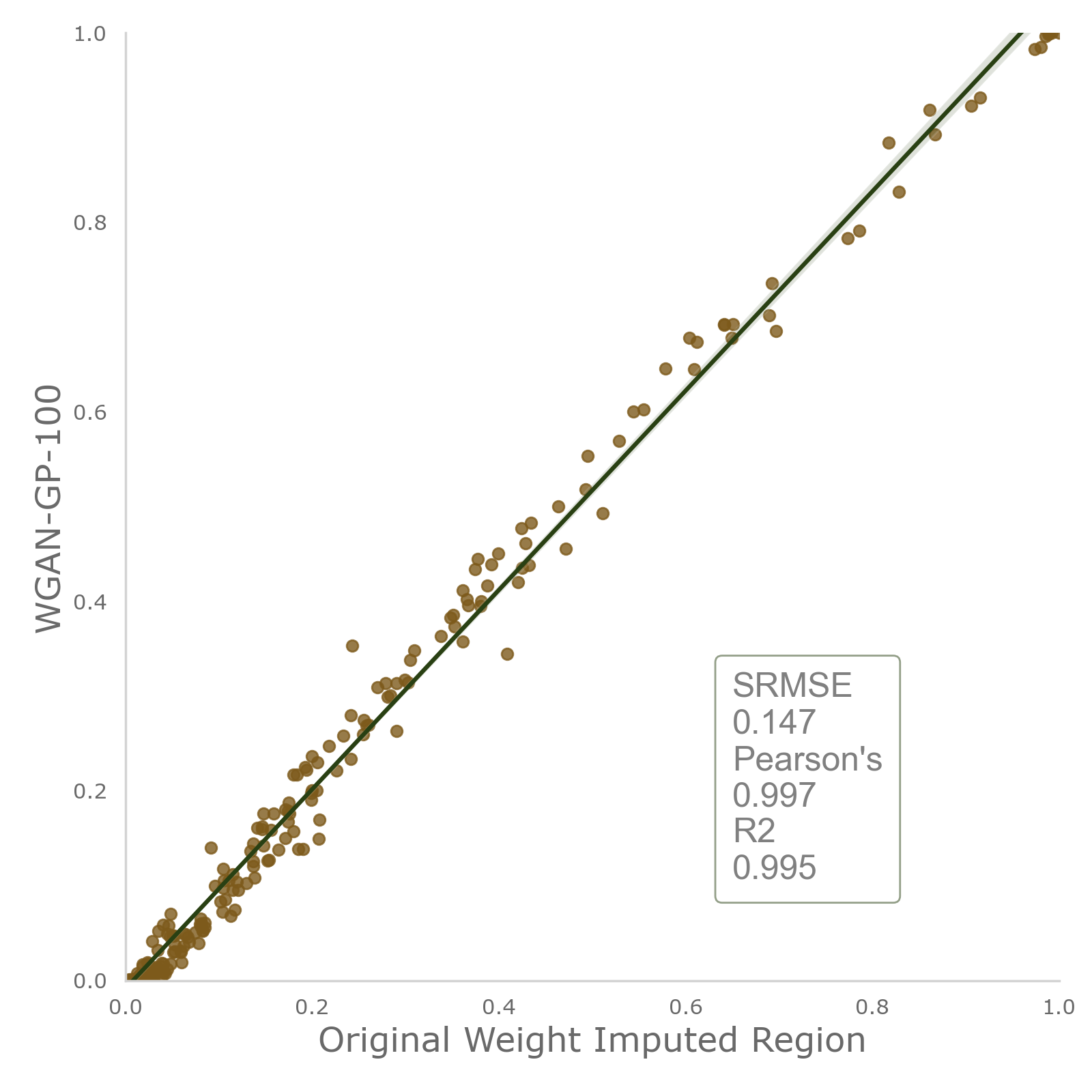}
    \caption[Univariate representation of WGAN on EU-SILC Region NUTS-2 with Helsinki from training on weight-imputed region data only]{Helsinki 2}
    \label{fig:helsinki-weight-imputed-from-region}
    \end{subfigure}
    \begin{subfigure}{0.22\textwidth}
    \centering
    \includegraphics[width=\textwidth]{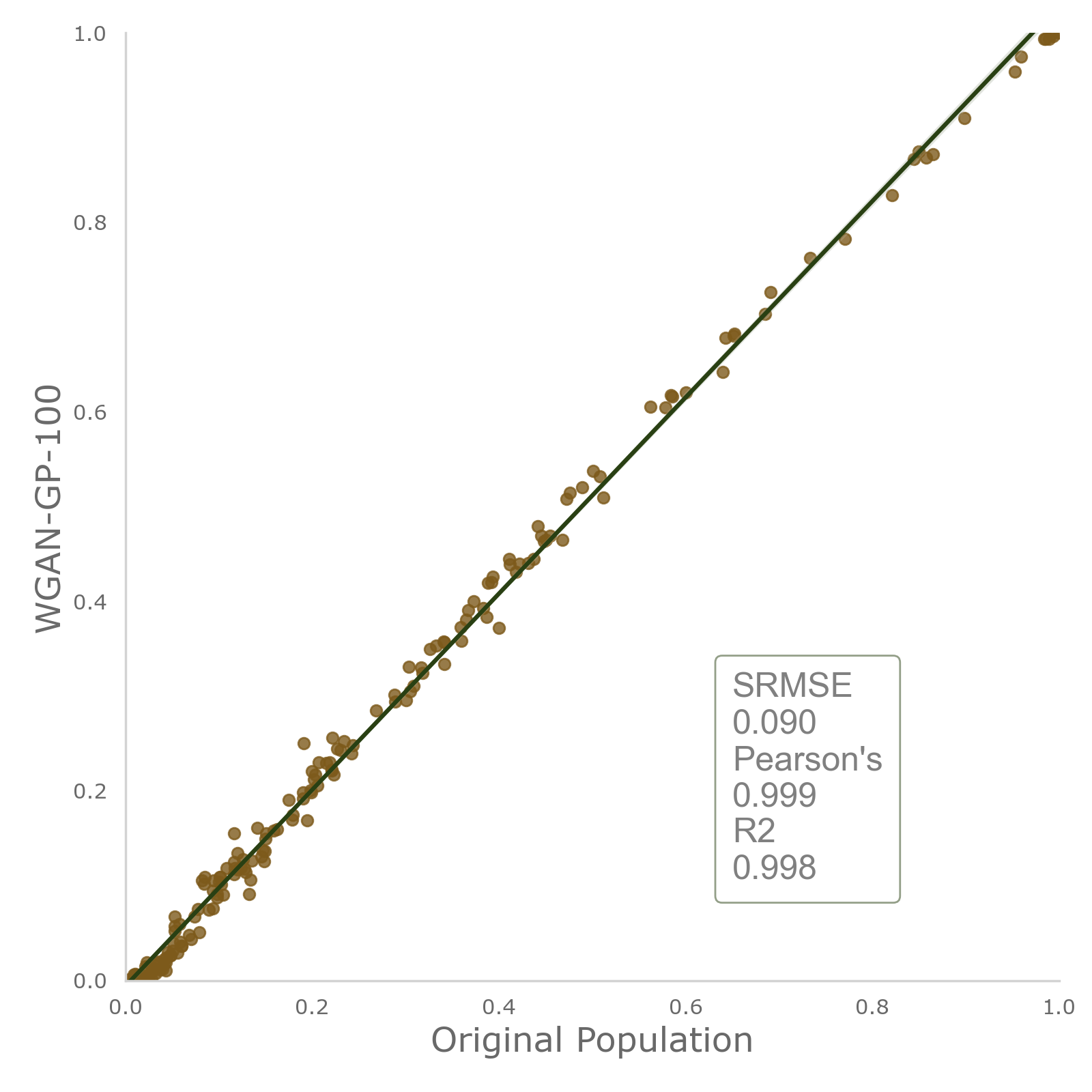}
    \caption[Univariate representation of WGAN on EU-SILC Region NUTS-2 with Helsinki from training on wgan-imputed region data only]{Helsinki 3}
    \label{fig:helsinki-wgan-imputed-from-region}
    \end{subfigure} 
\begin{subfigure}{0.22\textwidth}
    \centering
    \includegraphics[width=\textwidth]{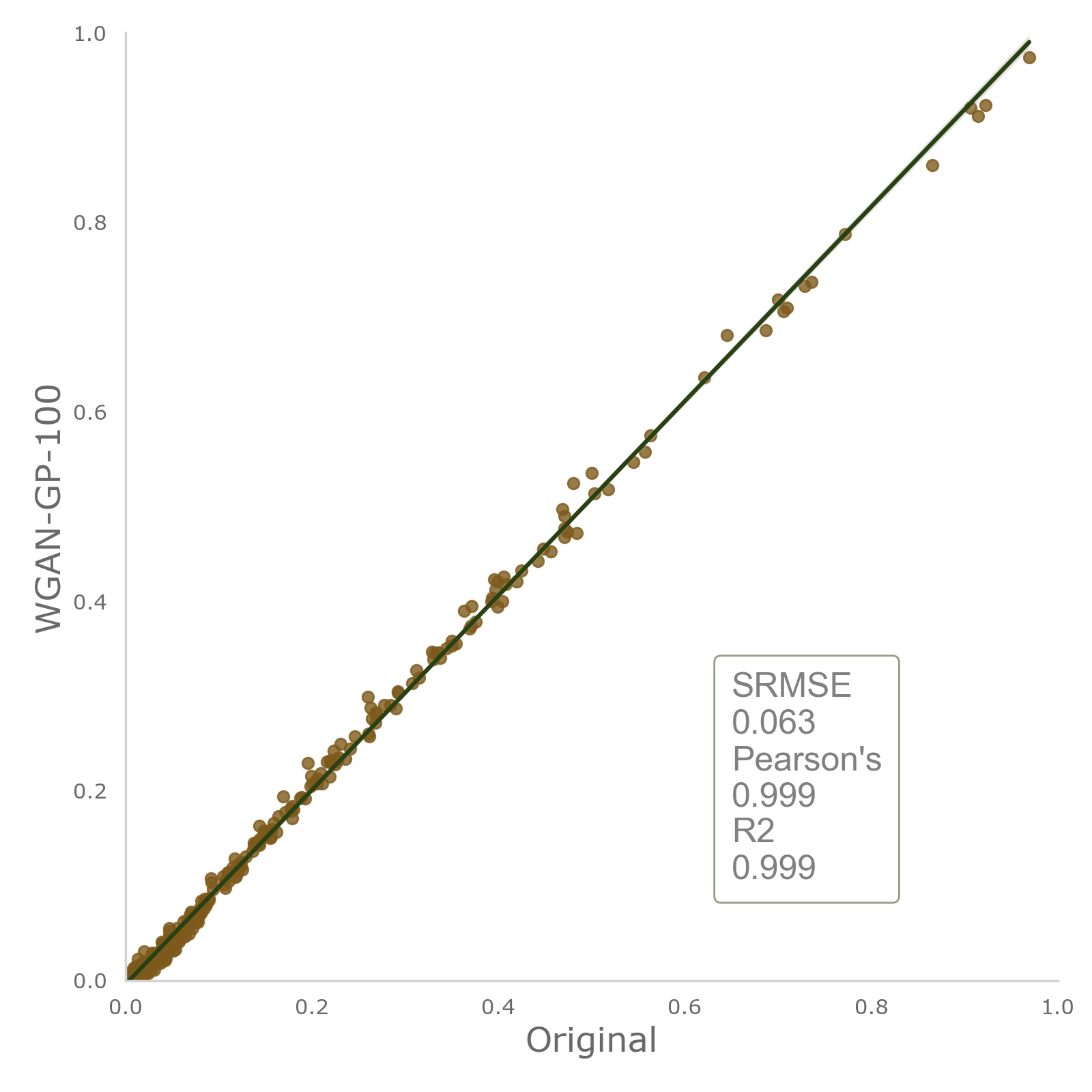}
    \caption[Univariate representation of WGAN on EU-SILC Greece Population]{Greece}
    \label{fig:greece-full-population}
    \end{subfigure}
\begin{subfigure}{0.22\textwidth}
    \centering
    \includegraphics[width=\textwidth]{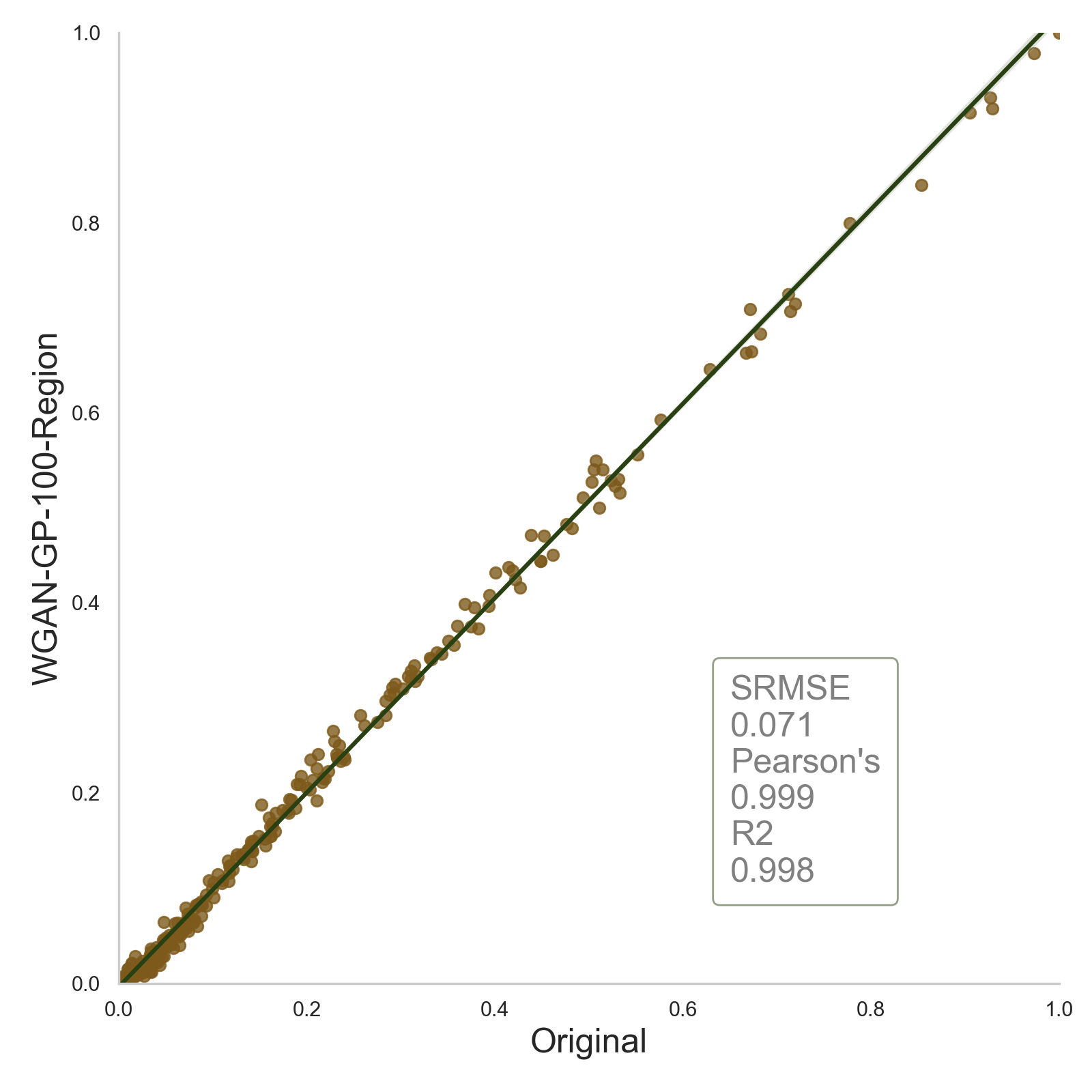}
\caption[Univariate representation of WGAN on EU-SILC Region NUTS-1 including Thessaloniki from training on complete national level data and extracting region]{Thessaloniki 1}
    \label{fig:thessaloniki-from-population}
    \end{subfigure}
\begin{subfigure}{0.22\textwidth}
    \centering
    \includegraphics[width=\textwidth]{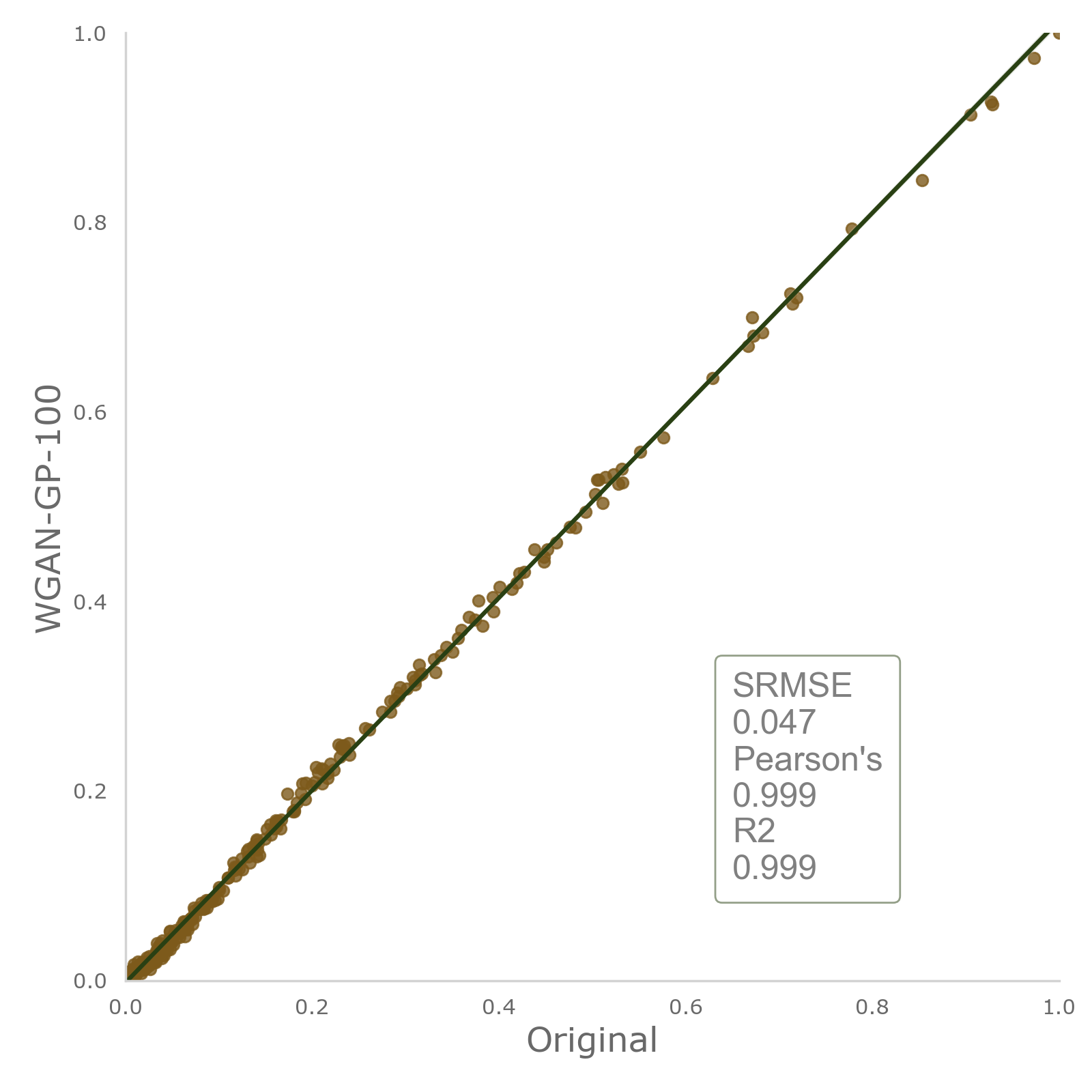}
    \caption[Univariate representation of WGAN on EU-SILC Region NUTS-1 with Thessaloniki from training on region data only]{Thessaloniki 2}
    \label{fig:thessaloniki-from-region}
    \end{subfigure}
\caption[Univariate Reproduction EU-SILC Finland]{Match between single variables in original and synthetic data from Wasserstein generative adversarial network. A comparison is made between variables from EU-SILC Finland in 2022. Figures a) and b) are produced training on weight-balanced complete population data. Figure c) is produced by training on the weight-balanced region, including Helsinki. Figure d) is produced by training on the wgan-imputed region, including Helsinki. Figures e) and f) are produced training on weight-balanced complete population data. Figure g) is produced by training on the weight-balanced region, including Thessaloniki only.}
  \label{fig:compare-synthetic-models-wgan-finland-greece}
\end{figure}

Balancing EU-SILC for Finland by weights, motivated by computational restrictions, is reduced by approximately 30-fold, giving an integer range for each original data record of 1 to 260. The dataset is upscaled from the original 17515 to 104668, of which 30509 belongs to the region where Helsinki is. Performance of the duplication by weights on Finland is shown in figure \ref{fig:compare-synthetic-models-wgan-finland-greece} and \ref{fig:bland-altmann-region-from-population-Finland}.

Balancing the Finnish data by WGAN-imputing according to the aggregated statistics, without removing any original data records, resulted in an up scaled training data set from 17515 to 65666 for the country-level data set.

\begin{figure}
    \centering
\begin{subfigure}{0.44\textwidth}
    \centering
    \includegraphics[width=\textwidth]{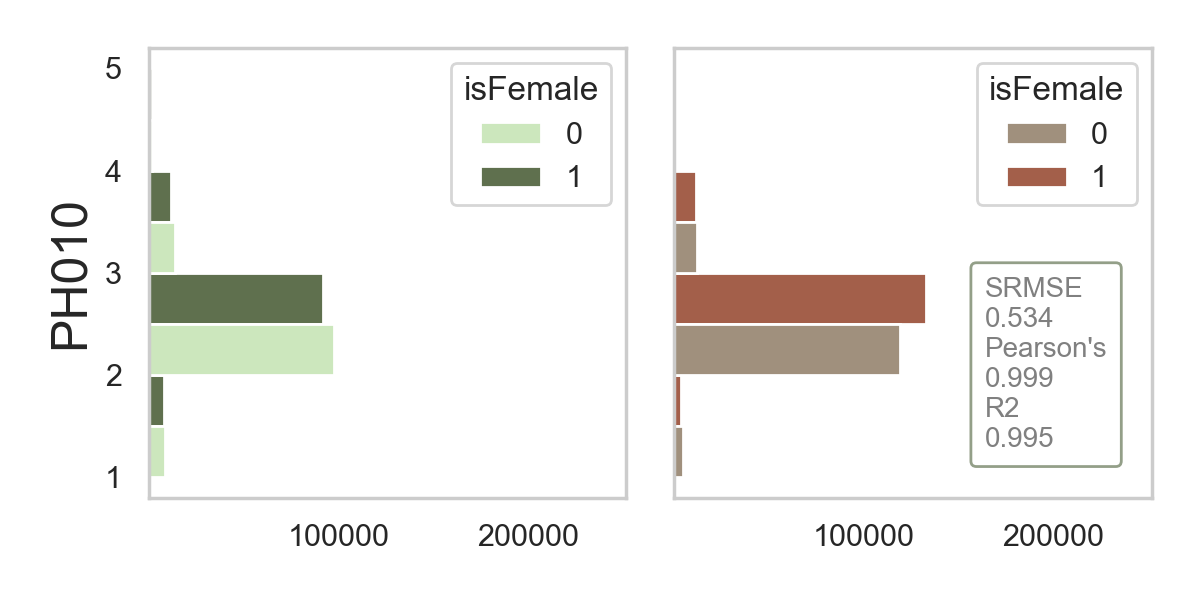}
    \caption[]{Imputed by weights}
    \end{subfigure}
\begin{subfigure}{0.44\textwidth}
    \centering
    \includegraphics[width=\textwidth]{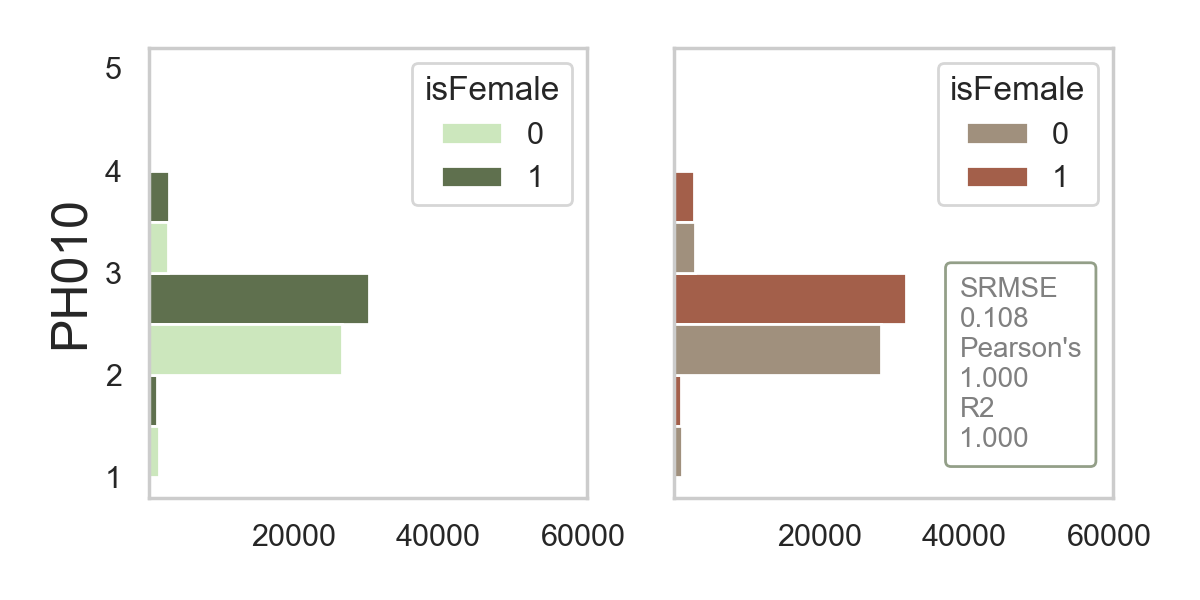}
    \caption[]{Imputed by demographic keys}
    \end{subfigure}
\caption[Self-perceived health (PH010)]{Reproduction of self-perceived health in synthetic populations trained on weight- and wgan-imputed original data to their respective training data.}
  \label{fig:PH010-Helsinki}
\end{figure}

\begin{figure}
    \centering
\begin{subfigure}{0.44\textwidth}
    \centering
    \includegraphics[width=\textwidth]{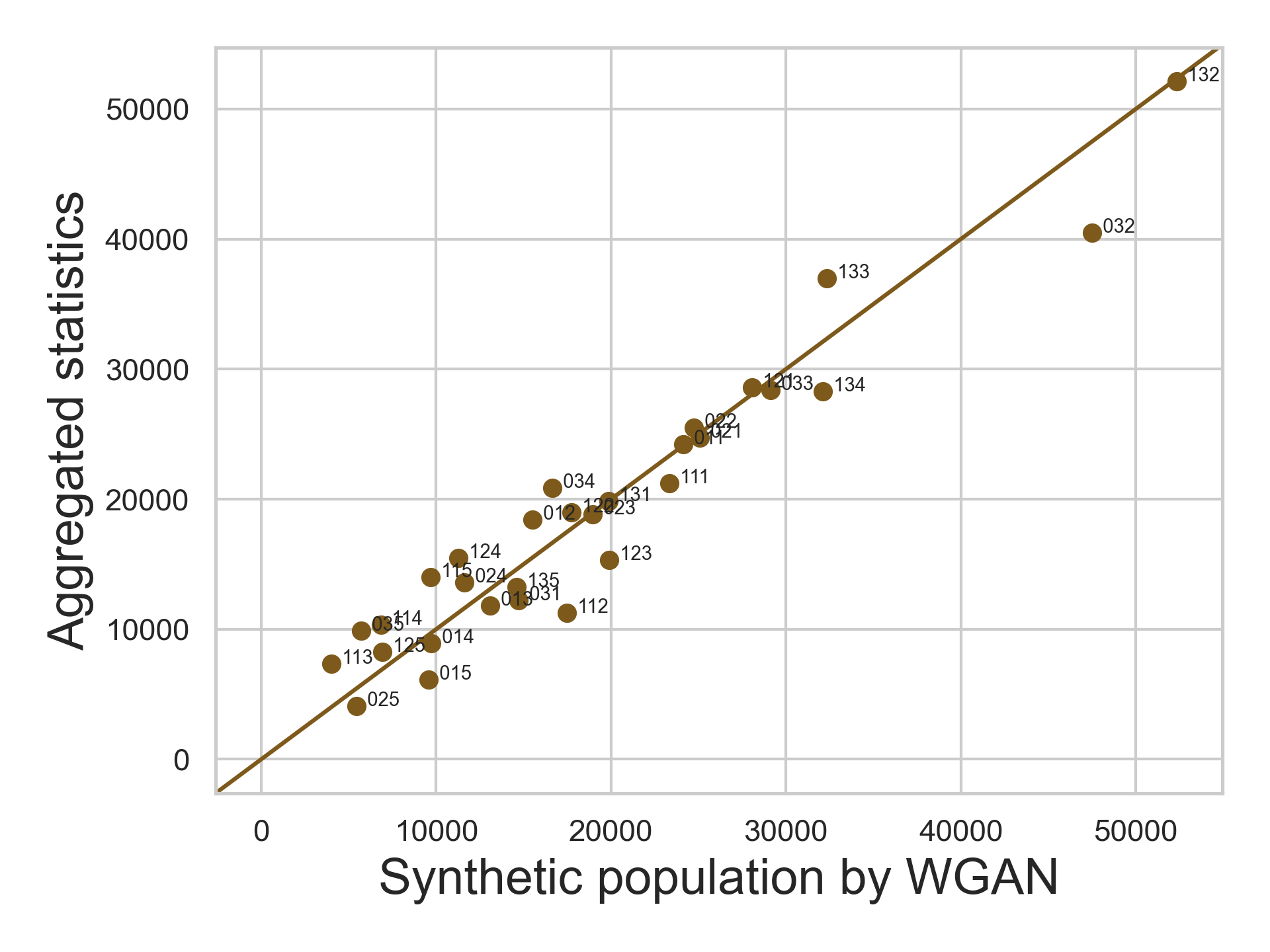}
    \caption[]{}
    \end{subfigure}
\begin{subfigure}{0.44\textwidth}
    \centering
    \includegraphics[width=\textwidth]{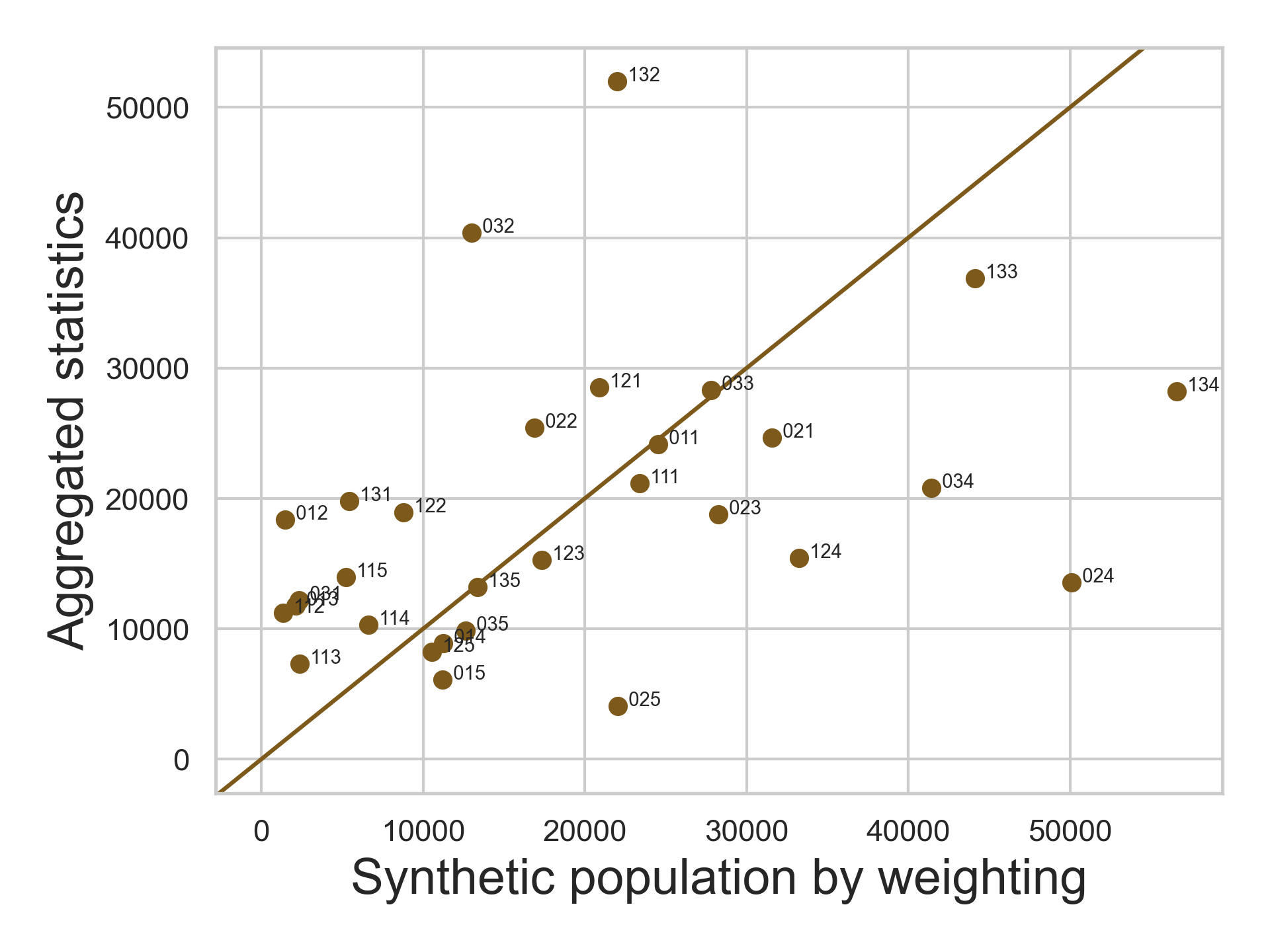}
    \caption[]{}
    \end{subfigure}
\caption[Helsinki demographic keys to synthetic population]{Comparison of aggregated statistics on gender, education and age to the final full-scale synthetic population of Helsinki created from the population of Finland and balanced by WGAN synthetic data (Figure a). The fit between aggregated statistics and synthetic population trained on weight-imputed original data is shown in Figure b.}
  \label{fig:Aggregated-keys-Helsinki}
\end{figure}

\subsection{Thessaloniki}
\begin{figure}
    \centering
     \begin{subfigure}{0.30\textwidth}
    \centering
    \includegraphics[width=\textwidth]{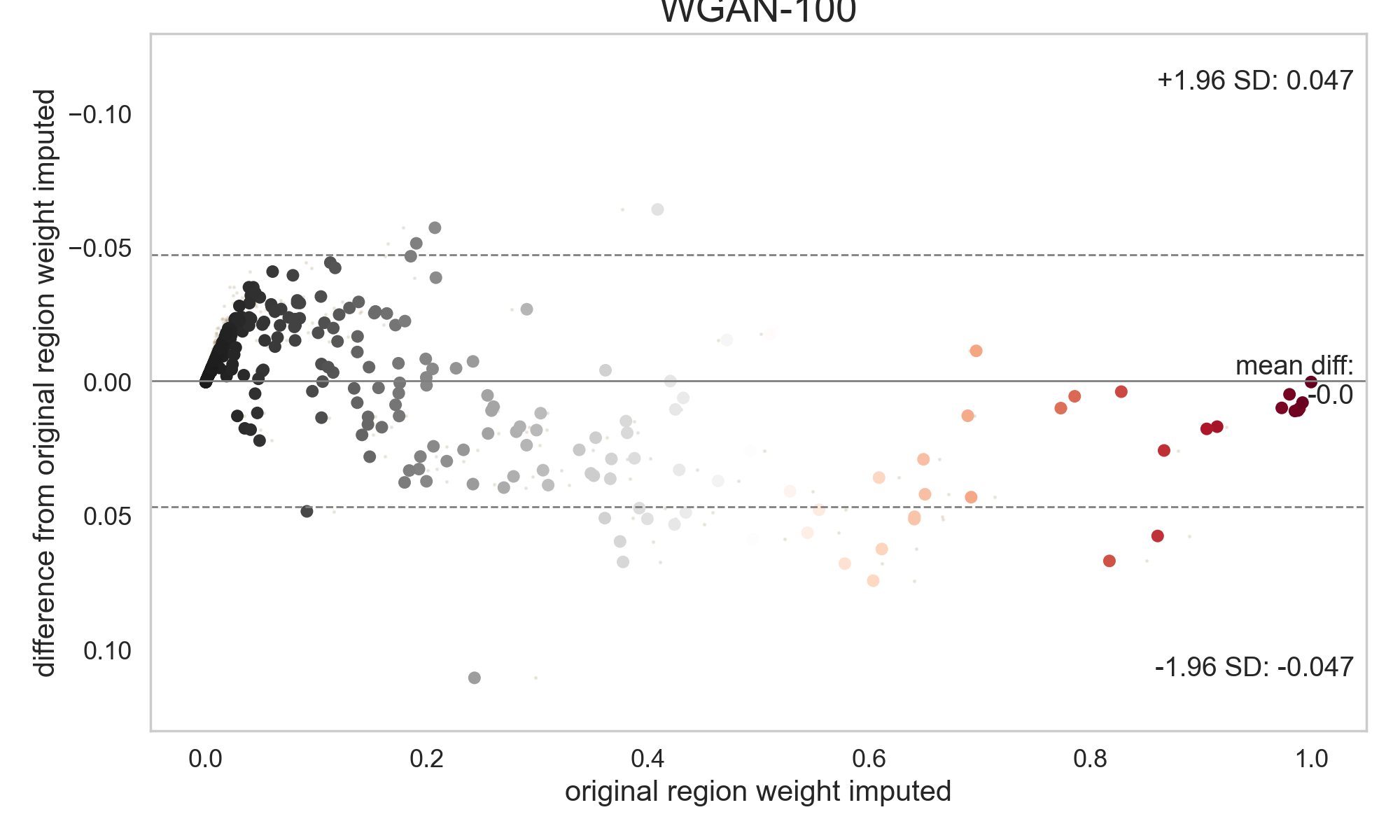}
    \caption[Bland-Altman for weight-imputed synthetic population Finland]{Helsinki (weight)}
    \label{fig:bland-altmann-region-from-population-Finland}
    \end{subfigure}
         \begin{subfigure}{0.30\textwidth}
    \centering
    \includegraphics[width=\textwidth]{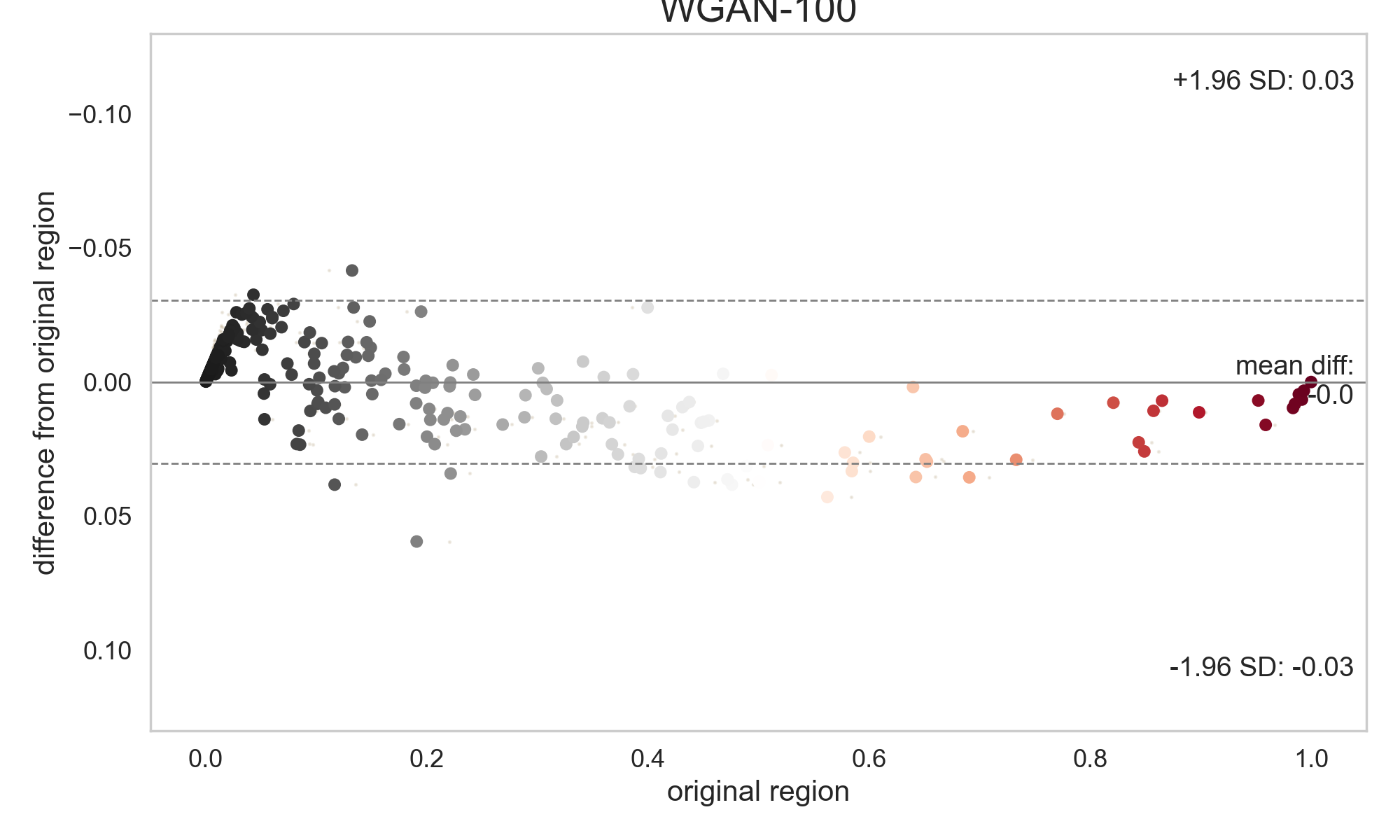}
    \caption[Bland-Altman for wgan-imputed population Helsinki]{Helsinki (wgan)}
    \label{fig:bland-altmann-region-from-wgan-imputed-region-Finland}
    \end{subfigure} 
         \begin{subfigure}{0.30\textwidth}
    \centering
    \includegraphics[width=\textwidth]{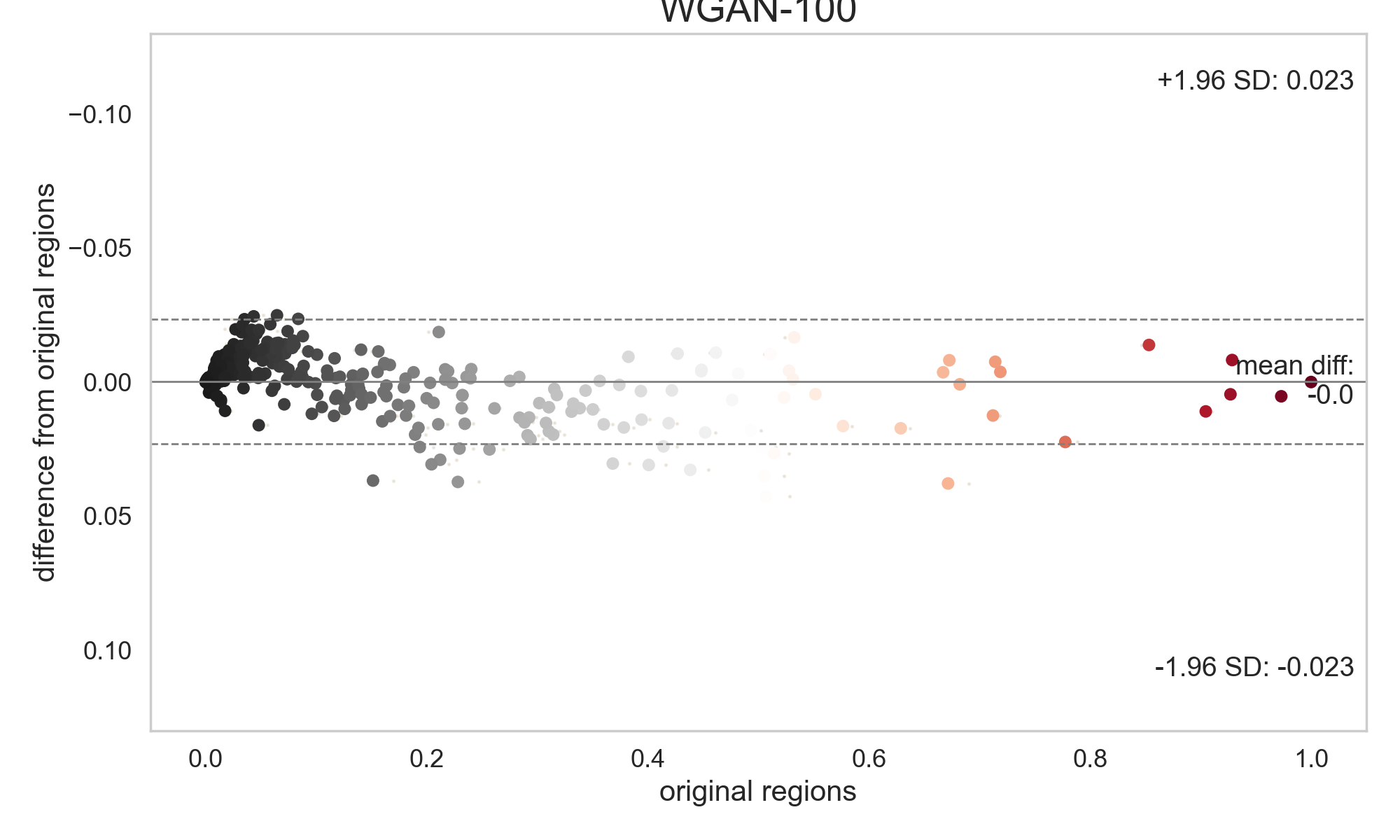}
    \caption[Bland-Altman for population in Thessaloniki]{Thessaloniki (weight)}
    \label{fig:bland-altmann-region-from-population-Greece}
    \end{subfigure}  
\caption[Bland-Altman for the region with Helsinki and Thessaloniki from EU-SILC]{Bland-Altman plots. a) Plot for synthetic data trained on weight-imputed originals for Helsinki municipality. b) Plot for synthetic data trained on wgan-imputed originals for Helsinki. c) Plot for the region to which Thessaloniki municipality belongs.
The points outside the two confidence interval lines are variable-value combinations that would analytically measure significantly different from the original data.}
  \label{fig:bland-altman-Finland-Greece}
\end{figure}

A demographic key on gender, age, and education was not publicly available to Thessaloniki. Weight imputing is, therefore, the only option for balancing data from Greece. The duplicating results in a data set of 308559 records, of which 108075 belong to the NUTS-1 region in which Thessaloniki belongs. Within the computational limits of a laptop, a country-level synthetic population above 700000 was impossible to produce. The maximum generated national population resulted in a slightly lower than actual synthetic Thessaloniki city population at age 16 or above of 219899 individuals. This synthetic population is an extract from the NUTS-1 region where Thessaloniki city is situated, not the city only, which is not directly available from the data. The performance of the synthetic population for Greece is shown in Figure \ref{fig:compare-synthetic-models-wgan-finland-greece} and Figure \ref{fig:bland-altmann-region-from-population-Greece}.

\subsection{Effects on Self-Perceived Health}

The synthetic populations for Helsinki are trained on both weight- and wgan-imputed original data, while for Thessaloniki, only weight-imputed are used. The wgan-impute are based on aggregated statistics. However, the wgan-imputed training data are by default contaminated by the influence of the Wasserstein Generative Adversarial Network. The hit on aggregated statistics is better for the wgan-imputed synthetic population for Helsinki \ref{fig:Aggregated-keys-Helsinki}. The Bland-Altman plot shows narrower confidence intervals and fewer outliers for the wgan-imputed synthetic population \ref{fig:bland-altmann-region-from-wgan-imputed-region-Finland} than the weight-imputed synthetic population \ref{fig:bland-altmann-region-from-population-Finland}. 

The reconstruction of self-perceived health (PH010) for Helsinki's weight- and wgan-imputed synthetic population are measured on the duplicate demographic keys of gender, age and education \ref{fig:PH010-skew-Helsinki}. Figure \ref{fig:PH010-wgan-to-wgan-imputed-original} compares the distribution of self-perceived health in the wgan-imputed synthetic population to wgan-imputed original training data. The fit seems good. However, when comparing the same synthetic population to the weight-imputed original on the duplicate demographic keys, the over-representation of the most frequent value (PH010=2) and under-representation of less frequent values show up \ref{fig:PH010-wgan-to-weight-imputed-original}. The synthetic population based on weight-imputed compared to the weight-imputed original shows much more spread but does not express a distorted pattern.

%% file: 4-Discussion/discussion.tex
\section{Discussion}
\label{sec:Discussion}


Deep generative methods are currently the only computational tractable methods to produce a population with numerous attributes from original microdata. As the number of attributes increases, it becomes difficult to ensure the preservation of statistical relationships between variables in the original data. Original microdata usually need to be balanced, and the balancing processes add to the challenges. No standardised measures exist for evaluating synthetic populations in general and high-attributed synthetic populations in particular. New techniques like neural manifold and diffusion probabilistic methods \cite{Manifold-LeCun-2022} show potential in better-preserving data patterns in the area of imaging from which the Wasserstein Generative Adversarial Networks derive. Further research is needed to enhance statistical, internal, and external validity for synthetic populations. 

The replication of tabular data, like the EU-SILC datasets, by Wasserstein Generative Adversarial Networks for all synthetic populations in this project are excellent, measured by Pearson's correlation coefficient, R-squared and SRMSE as shown in Figure \ref{fig:compare-synthetic-models-wgan-finland-greece}. The Bland-Altman plots in Figure \ref{fig:bland-altman-Finland-Greece} show that some variables might fail to substitute their originals in all the populations generated in this study. There seem to be slight differences in the quality between using country-level data or just extracting original data for the region. As some of the patterns in data would train better with more data, it is premature to conclude on this matter. 

The distortion in the variable self-perceived health (PH010) (Figure \ref{fig:PH010-skew-Helsinki}) raises particular challenges of discrimination. The double effect of first wgan-imputing (Approach 1) and then training a second wgan on these data poses particular challenges for representing persons with fringe values on essential variables and discrimination, highlighting the urgency and complexity of the issue. The distortion effect is hidden when comparing the synthetic population to its wgan-imputed original, as shown in Figure \ref{fig:PH010-wgan-to-wgan-imputed-original}. The weight-imputing (Approach 2) strategy seems a better fit for dealing with less distortion of a variable like self-perceived health.

At a country level, the EU-SILC data have a floating number person weight variable to balance the data to ensure proper representation of all groups. To create representative synthetic populations at a more granular level, like the Helsinki and Thessaloniki municipalities, two approaches to balancing are explored: the use of demographic profiles by wgan-replicas (Approach 1) and weight-imputing (Approach 2) to fit the municipalities. Not surprisingly, the best fit to demographic profiles is the strategy using wgan-replicas (Approach 1) as shown in Figure \ref{fig:Aggregated-keys-Helsinki}. Weight-imputing on a regional level does not guarantee an excellent fit to a simple demographic key like gender, age and education (Figure \ref{fig:Aggregated-keys-Helsinki}).

%% file: 5-Conclusion/conclusion.tex
\section{Conclusion}
\label{sec:Conclusion}

This work demonstrates the promising potential of Wasserstein generative adversarial networks (WGAN) in producing high-attributed synthetic populations from EU-SILC data, as measured by standardised root mean squared error (SRMSE), Pearson's correlation coefficient, and R-squared. These populations are suitable for use in simulations like agent-based modelling. The variable fit between the original data and the synthetic population with many features can be visually validated in Bland-Altman plots, further underlining the potential of WGANs.

The limitations of WGANs and other deep generative methods used on EU-SILC data include the tendency to under-represent fringe profiles represented by the variable self-perceived health, which raises questions of discrimination \ref{fig:PH010-skew-Helsinki}. In this material, the weight-imputed training data seems less likely to skew self-perceived health than the wgan-imputed. On the other hand, training with wgan-imputed data provides a better fit for the demographic profile of Helsinki.

Validating synthetic populations is a complex task; no standard method is currently available. The complexity increases when dealing with high-featured populations. Future work should explore Bland-Altman and other methods to handle high-featured data, such as the neural manifold clustering techniques \cite{Manifold-LeCun-2022} used in imaging to preserve deeper statistical structures. However, the urgent need for further investigation lies in representing fringe group profiles.

%% file: 6-Appendix/appendix.tex
\appendix
\newpage
\begin{subappendices}
\renewcommand{\thesection}{\Alph{section}}
\section{Code}\label{WGAN-Code}

\subsection{Wasserstein Generative Adversarial Network} 
\begin{verbatim}
"""
WGAN-GP
Generative adversarial networks for synthetic population generation
using Wasserstein and gradient penalty.
"""
import torch
import torch.nn as nn

class Critic(nn.Module):
    """
    Class Critic is the neural network performing the
    critic functions in the Wasserstein Generative Adversarial
    Network.
    """
    def __init__(self, feature_dimension, output_dim=1):
        super(Critic, self).__init__()
        self.feature_dimension = feature_dimension
        self.critic = nn.Sequential(
            self._block(self.feature_dimension, 100),
            self._block(100, 150),
            nn.Linear(in_features=150, out_features=output_dim),
        )

    def _block(self, input_d, n_nodes):
        return nn.Sequential(
            nn.Linear(in_features=input_d, out_features=n_nodes),
            # do not use batch-norm in critic
            nn.InstanceNorm1d(n_nodes),
            nn.LeakyReLU(0.2))

    def forward(self, x):
        return self.critic(x)

class Generator(nn.Module):
    """
    Class Generator is a neural network performing the
    generative function in the Wasserstein Generative Adversarial
    Network.
    """
    def __init__(self, feature_dimension, latent_dimension):
        super(Generator, self).__init__()
        self.latent_dimension = latent_dimension
        self.feature_dimension = feature_dimension
        self.generator = nn.Sequential(self._block(self.latent_dimension, 150),
                                       self._block(150, 100),
                                       nn.Linear(100, self.feature_dimension),
                                       nn.Sigmoid())

    def forward(self, x):
        return self.generator(x)

    def _block(self, input_d, n_nodes):
        return nn.Sequential(
            nn.Linear(in_features=input_d, out_features=n_nodes),
            nn.BatchNorm1d(n_nodes),
            nn.LeakyReLU(0.2))

def initialise_weights(model):
    for m in model.modules():
        if isinstance(m, nn.Linear):
            nn.init.normal_(m.weight.data, 0.0, 0.02)

def gradient_penalty(model, real, fake, device="cpu"):
    batch_size = real.shape[0]
    feature_dimension = real.shape[1]
    # One epsilon per example
    epsilon = torch.rand(batch_size, 1).repeat(1, feature_dimension).to(device)
    interpolated = real * epsilon + fake * (1 - epsilon)
    mixed_score = model(interpolated)
    gradient = torch.autograd.grad(inputs=interpolated,
                                   outputs=mixed_score,
                                   grad_outputs=torch.ones_like(mixed_score),
                                   create_graph=True,
                                   retain_graph=True)[0]
    gradient = gradient.view(gradient.shape[0], -1)  # flatten
    gradient_norm = torch.linalg.vector_norm(gradient, ord=2, dim=1)
    gp = torch.mean((gradient_norm - 1) ** 2)
    return gp

\end{verbatim}

\section{Tools}
The WGAN code is written with PyTorch 2.1.2 using Python version 3.10 and provided in the appendix \ref{WGAN-Code}. The data are cleaned using Pandas version 2.3, applying Scikit-Learn 1.4.1 and Scipy 1.12.0 using Iterative-imputer and KNN-imputer to impute variables with no more than 50\% missing values for Greece and no more than 67\% for Finland in order to approximately match the number of variables included. All models are run on an Ubuntu 22.04 LTS Linux ThinkStation P330 Thiny with NVIDIA Quadro P1000 graphic card using a Jupyter Notebook with Anaconda. Statsmodels version 0.14.1 is used to calculate statistics. 

\section{Data Preparation} \label{data-preparation}
If more than 50\%  is missing, the EU-SILC Greece variables are excluded. To approximately match the variables left in the EU-SILC Greece, the limit is set to 67\% for the EU-SILC Finland. Imputing many variables like this can threaten external validity as the original data can diverge from the actual population. However, this is not considered an issue when demonstrating high-featured population generation. All non-binary variables, including numerical ones, are transformed into categorical intervals and are represented as one-hot-encoded. Binary data are represented with one variable with values zero or one. The data fed to the neural network has no missing values and is a complete binary vector with one column for each variable-value pair.

\section{Model Training} \label{Model_Training}
\begin{figure}
    \centering
\begin{subfigure}{0.32\textwidth}
    \centering
    \includegraphics[width=\textwidth]{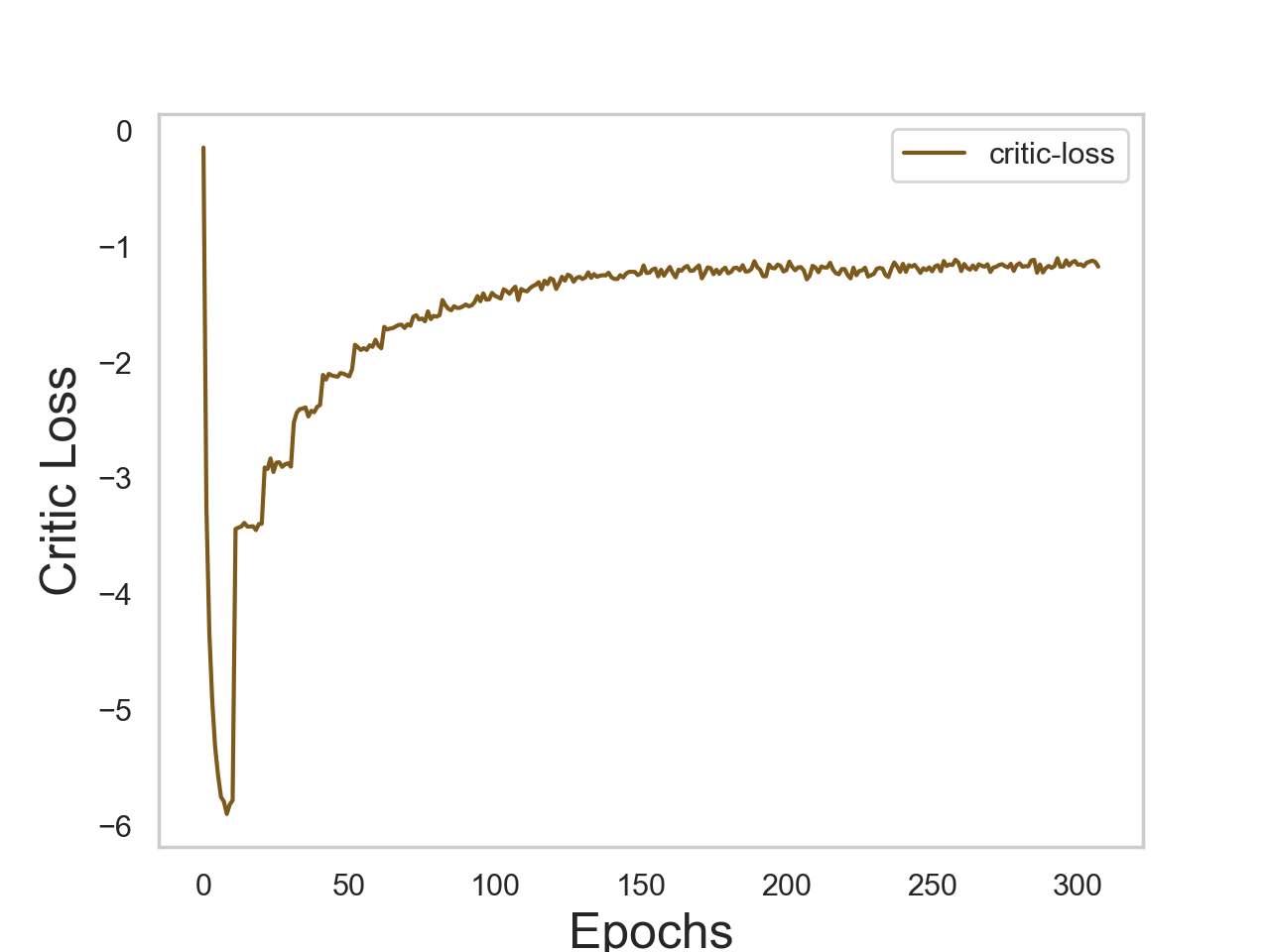}
    \caption[Critic Loss from Finland]{Critic Loss}
    \label{fig:critic_finland}
\end{subfigure}
\begin{subfigure}{0.32\textwidth}
    \centering
    \includegraphics[width=\textwidth]{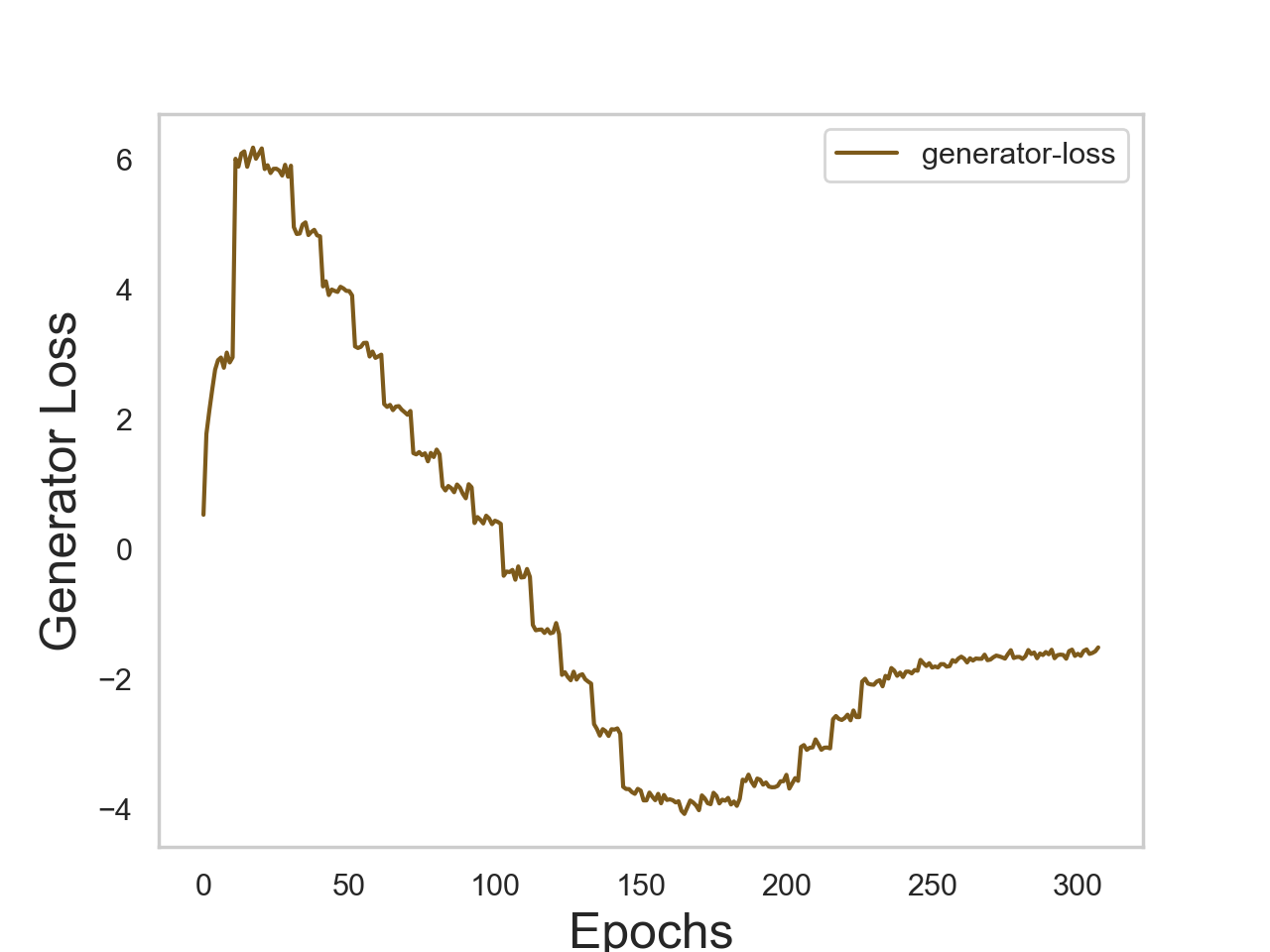}
    \caption[Generator Loss from Finland]{Generator Loss}
    \label{fig:generator_finland}
\end{subfigure}
\begin{subfigure}{0.32\textwidth}
    \centering
    \includegraphics[width=\textwidth]{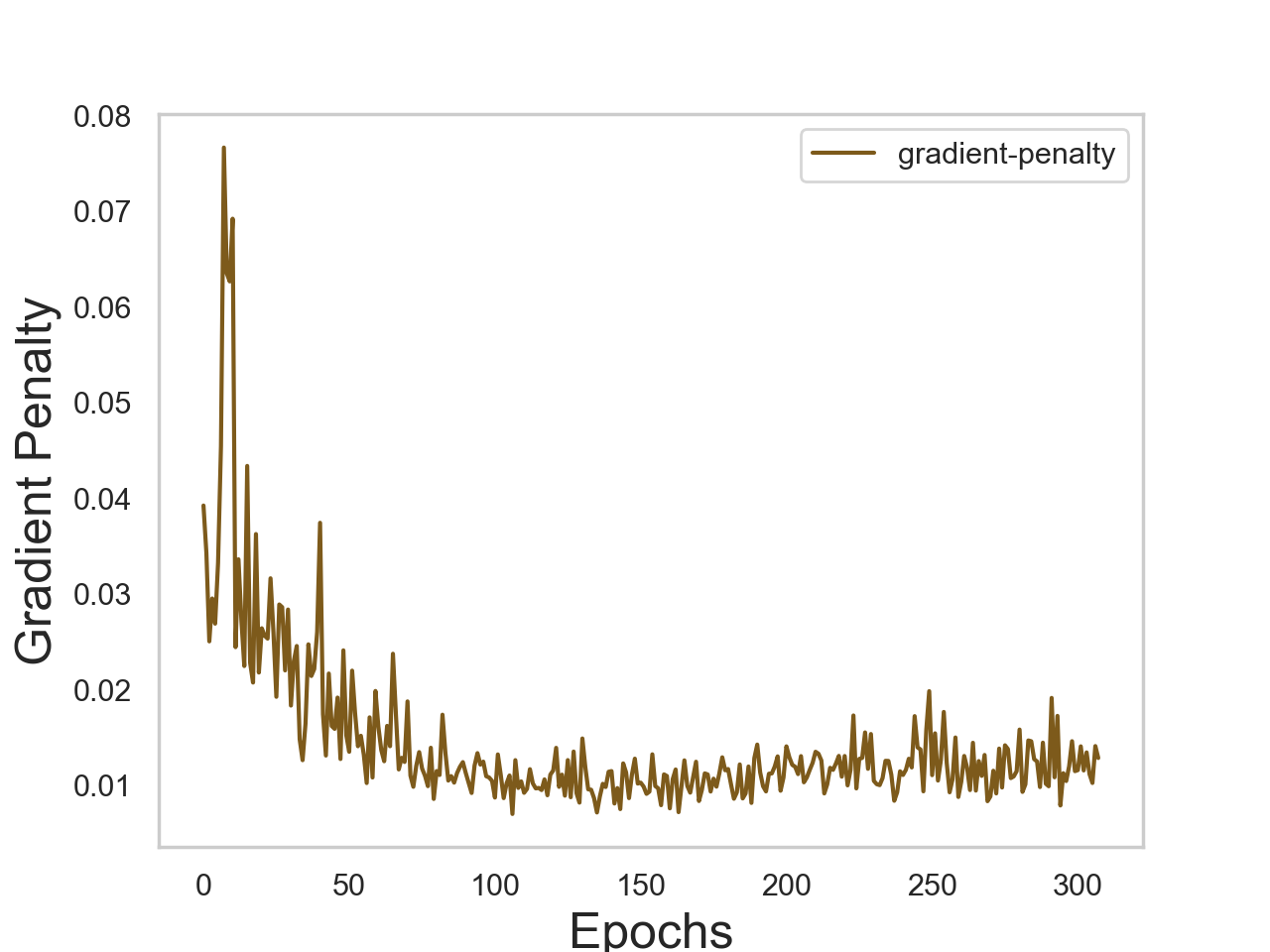}
    \caption[GP Loss from Finland]{GP Loss}
    \label{fig:gp_loss_finland}
\end{subfigure}
\begin{subfigure}{0.32\textwidth}
    \centering
    \includegraphics[width=\textwidth]{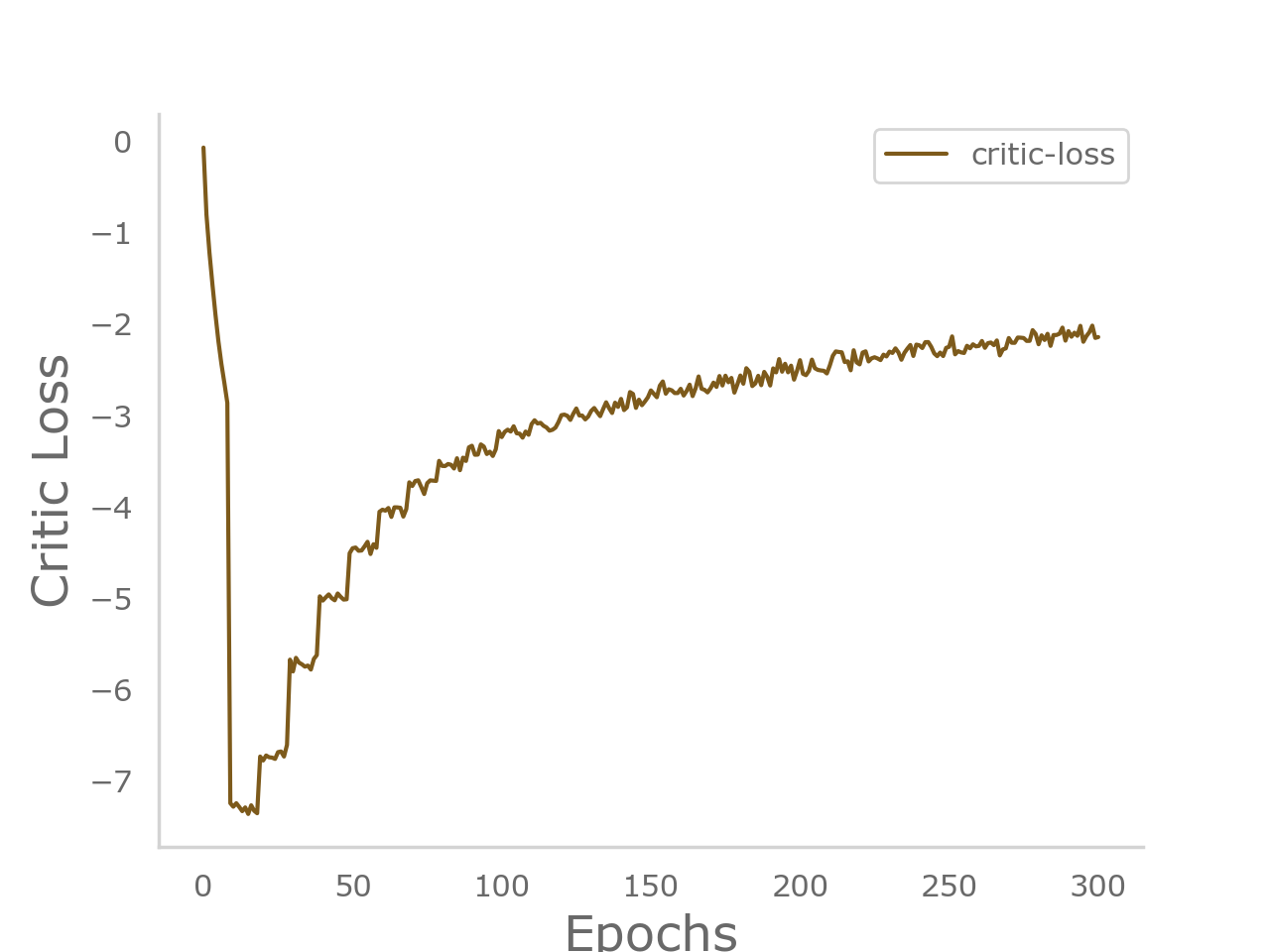}
    \caption[Critic Loss from Region Helsinki Finland]{Critic Loss}
    \label{fig:critic_region_finland}
\end{subfigure}
\begin{subfigure}{0.32\textwidth}
    \centering
    \includegraphics[width=\textwidth]{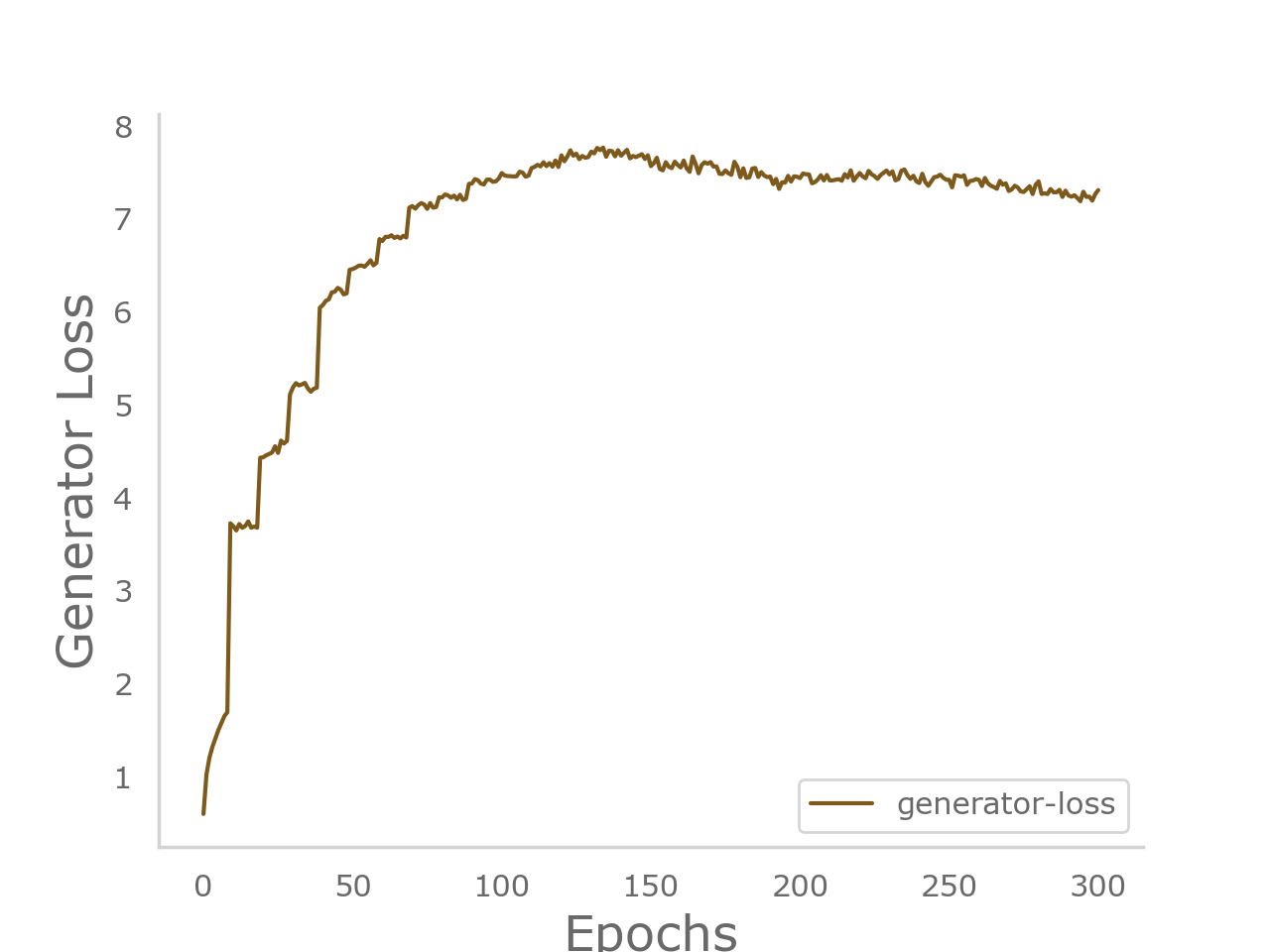}
    \caption[Generator Loss from Region Helsinki Finland]{Generator Loss}
    \label{fig:generator_region_finland}
\end{subfigure}
\begin{subfigure}{0.32\textwidth}
    \centering
    \includegraphics[width=\textwidth]{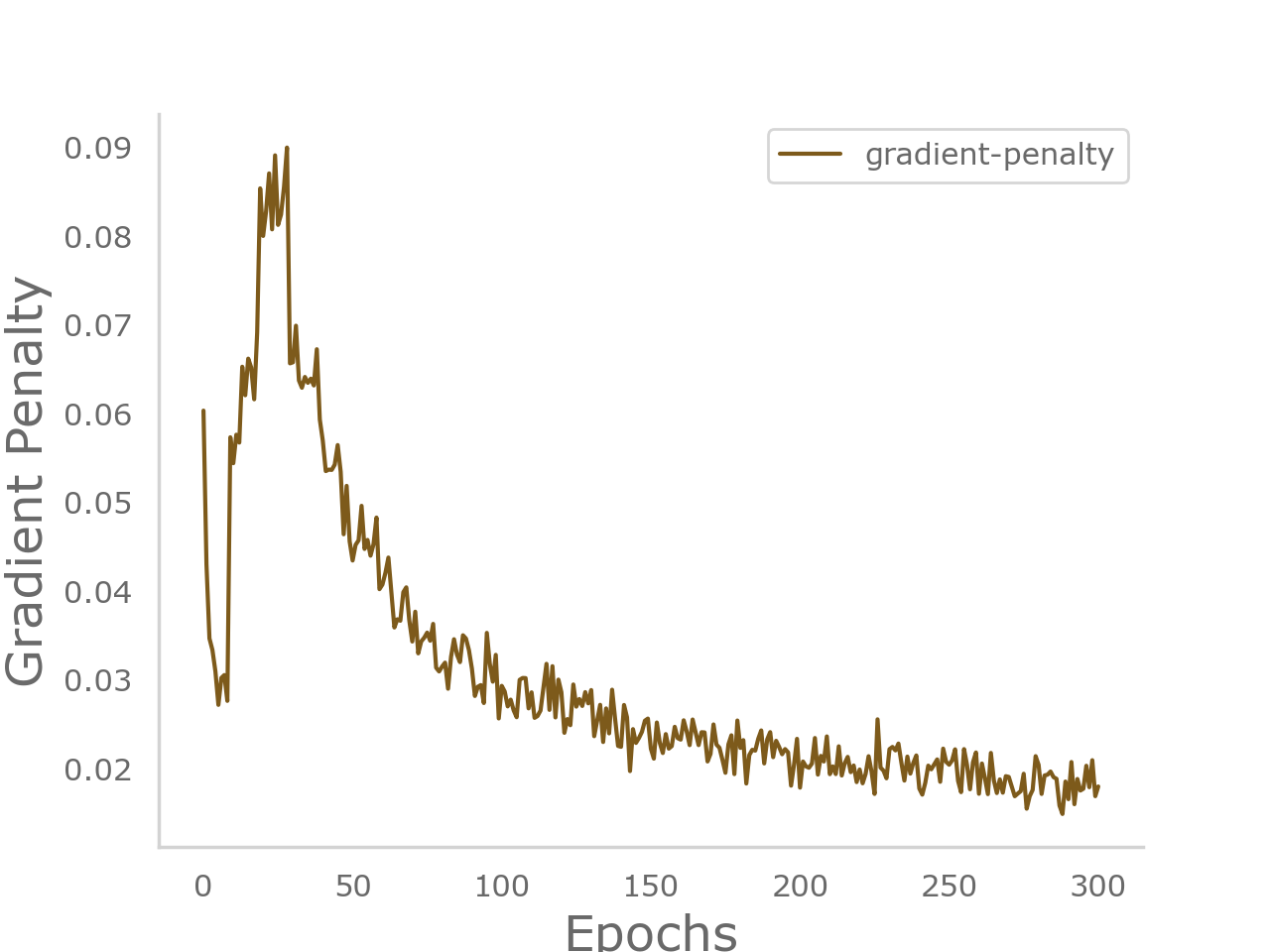}
    \caption[GP Loss from Region Helsinki Finland]{GP Loss}
    \label{fig:gp_loss_region_finland}
\end{subfigure}
\caption[Training curves for Wasserstein generative adversarial network on unweighted EU-SILC Finland]{Training curves for Wasserstein generative adversarial network on unweighted EU-SILC Finland (size 17515)(a, b, c). On weighted EU-SILC Finland (size 104668) (d, e, f). Models are run in 300 iterations with a learning rate of 0.00001 and batch size of 300. Adam optimiser and activation by leaky RELU. See further details in the code \ref{WGAN-Code}.}
  \label{fig:wgan-training-finland}
\end{figure}

\begin{figure}
    \centering
\begin{subfigure}{0.32\textwidth}
    \centering
    \includegraphics[width=\textwidth]{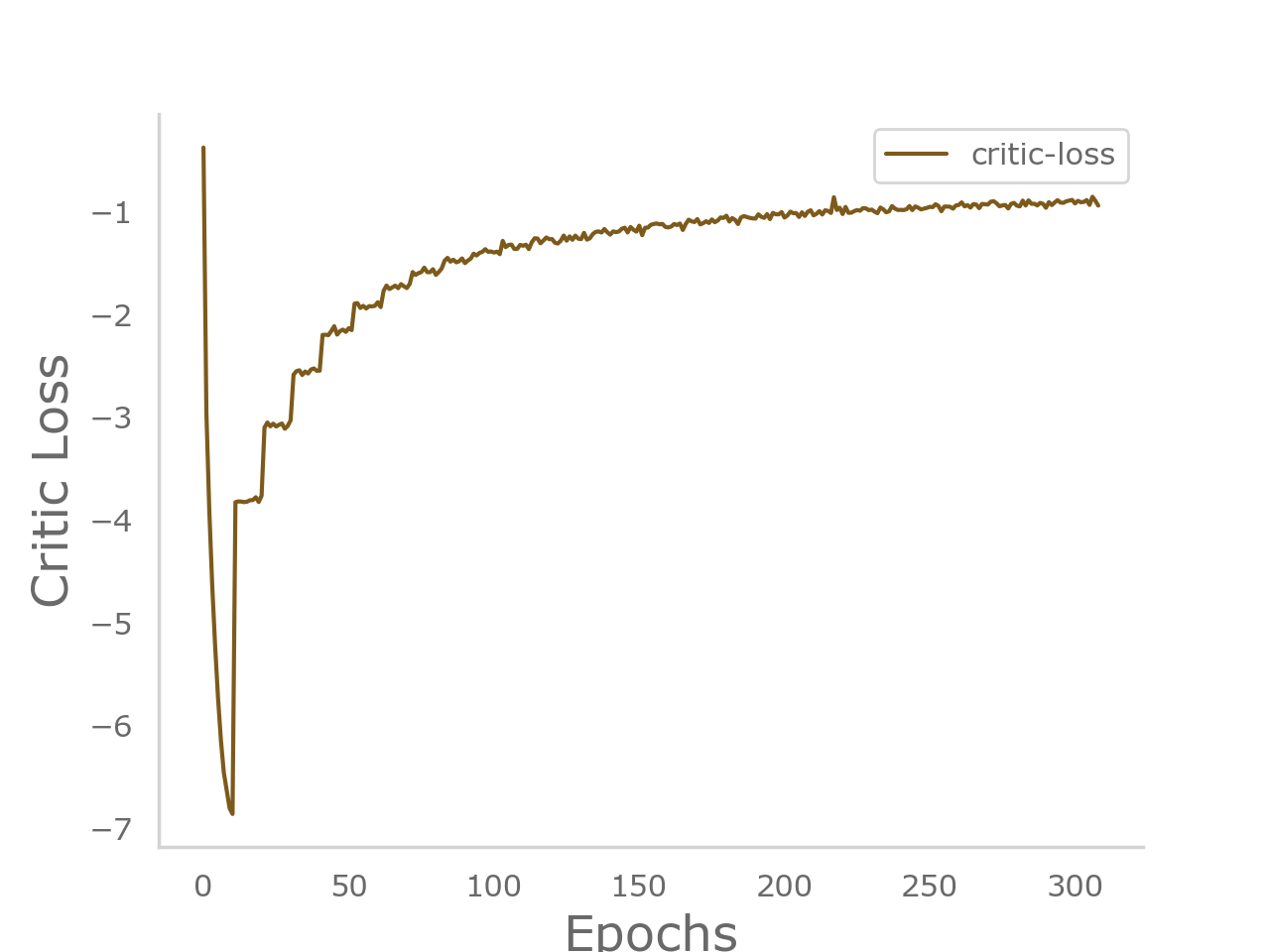}
    \caption[Critic Loss from Greece]{Critic Loss}
    \label{fig:critic_greece}
\end{subfigure}
\begin{subfigure}{0.32\textwidth}
    \centering
    \includegraphics[width=\textwidth]{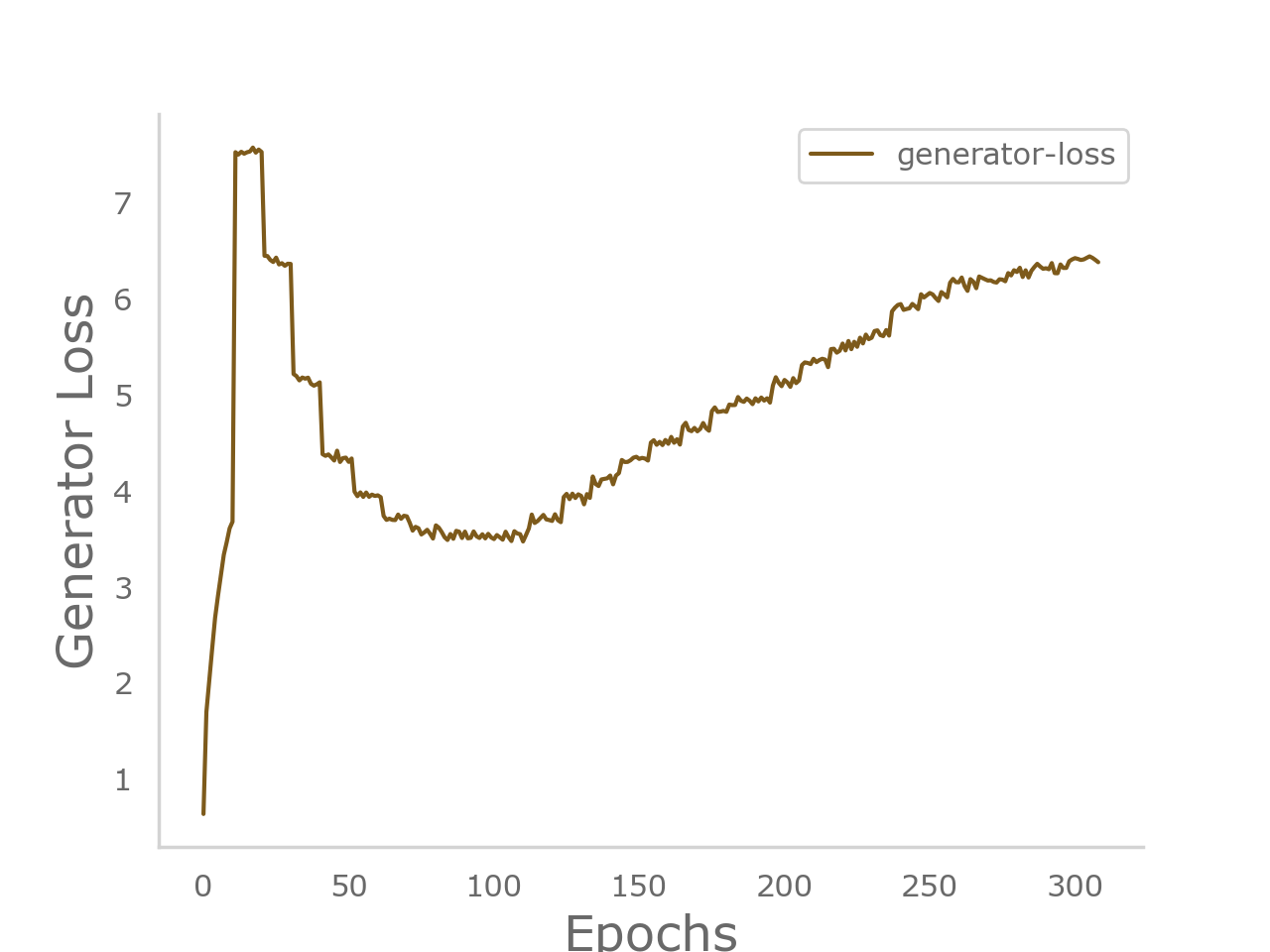}
    \caption[Generator Loss from Greece]{Generator Loss}
    \label{fig:generator_greece}
\end{subfigure}
\begin{subfigure}{0.32\textwidth}
    \centering
    \includegraphics[width=\textwidth]{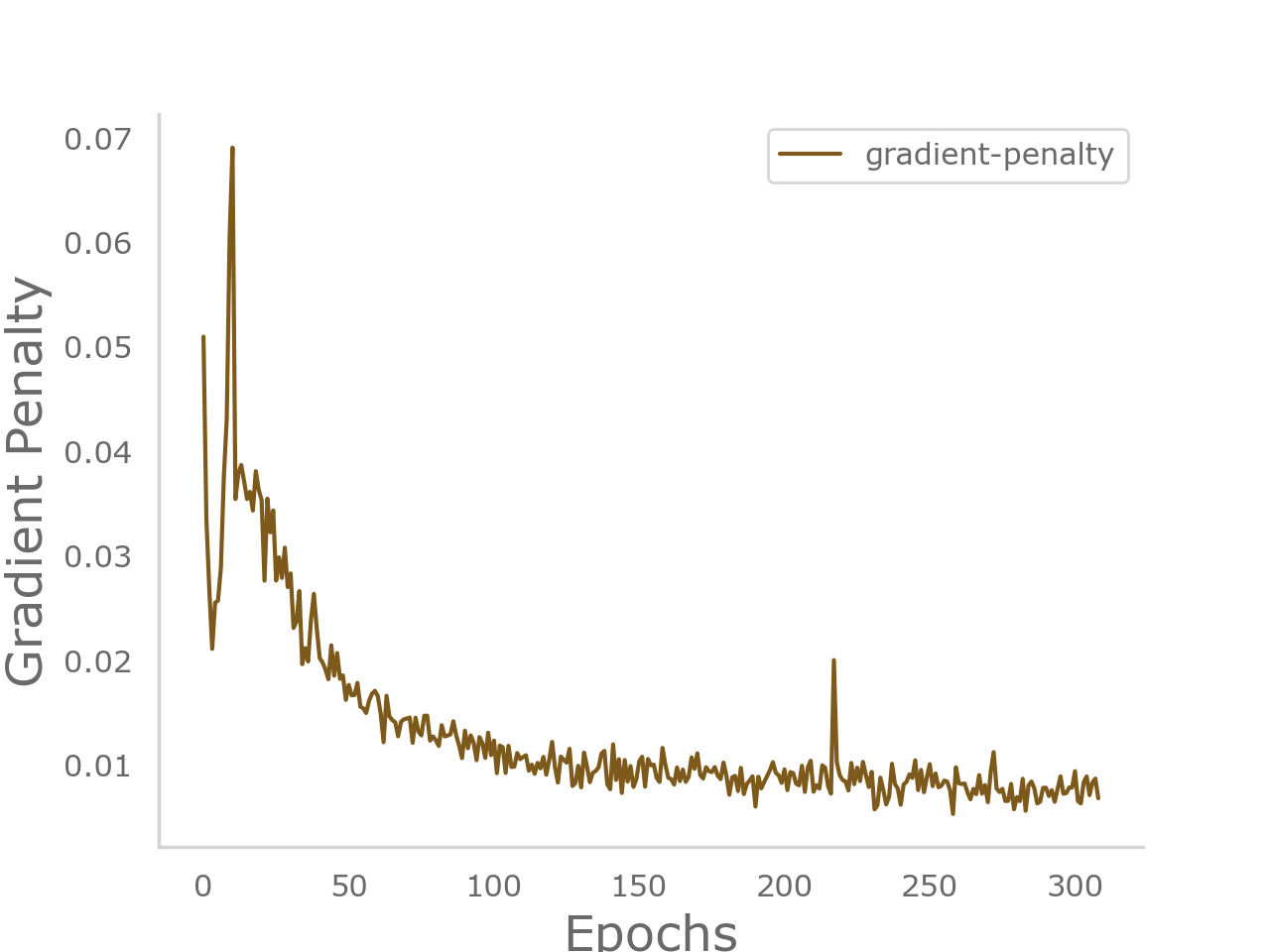}
    \caption[GP Loss from Greece]{GP Loss}
    \label{fig:gp_loss_greece}
\end{subfigure}
\caption[Training curves for Wasserstein generative adversarial network on weight imputed EU-SILC Greece]{Training curves for Wasserstein generative adversarial network on weight imputed EU-SILC Greece. The data was up-scaled from the original unweighted 19480 to 308559 weighted records by adding duplicate originals to match the weights. The feature size of one-hot-encoded and binary variables was 404. The learning rate was set to 0.00001, batch size 200, running 300 iterations.}
  \label{fig:wgan-training-greece}
\end{figure}

Overfitting neural network models will make them copy originals rather than generate new data records. The models are trained with a loss on the generator, the critic and the GP. The training is stopped just after convergence to avoid overfitting. See figure \ref{fig:wgan-training-finland} and \ref{fig:wgan-training-greece}. 

\begin{figure}
    \centering
     \begin{subfigure}{0.44\textwidth}
    \centering
    \includegraphics[width=\textwidth]{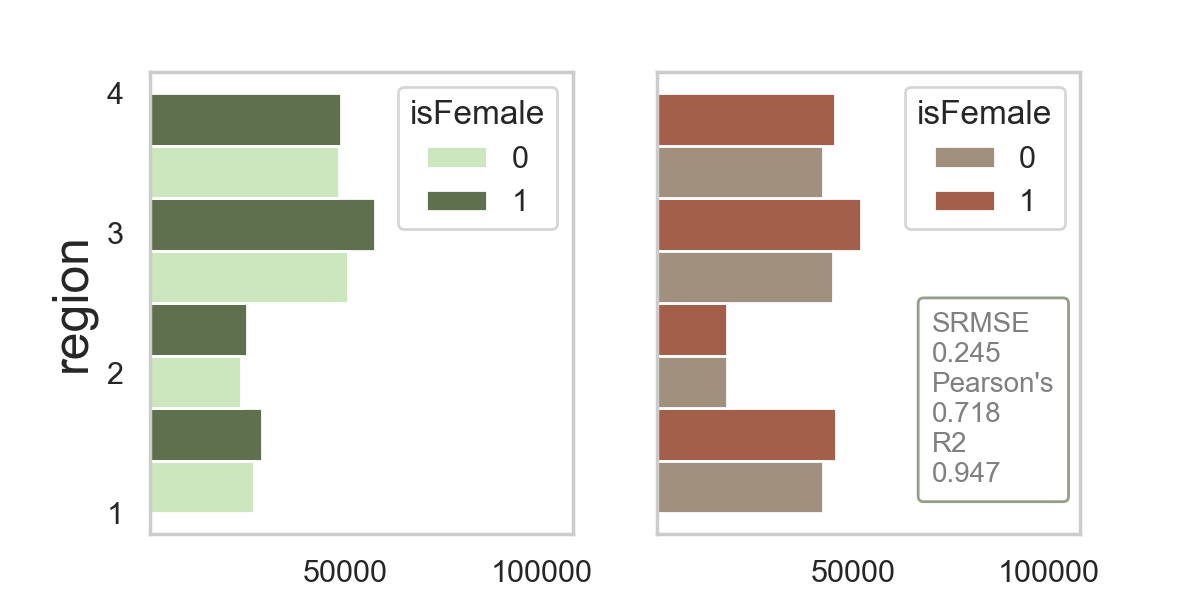}
    \caption[Reproduction of regions in Greece]{Regions in Greece}
    \label{fig:region-reproduction-Greece-sub}
    \end{subfigure}
\begin{subfigure}{0.44\textwidth}
    \centering
    \includegraphics[width=\textwidth]{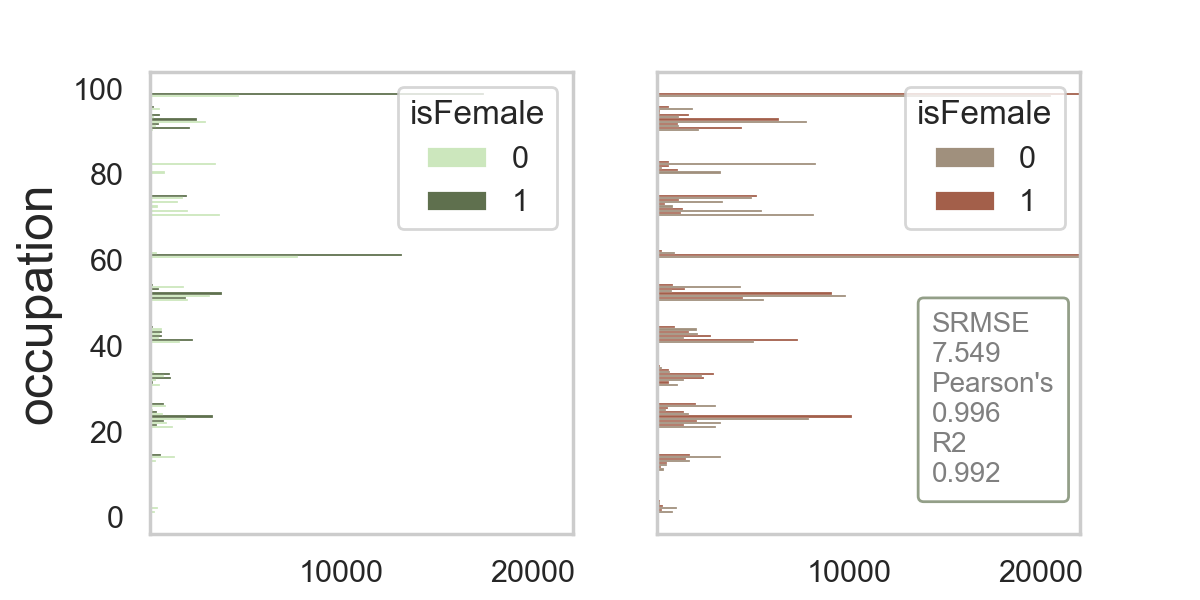}
    \caption[Reproduction of occupation in Greece]{Occupation in Greece}
    \label{fig:occupation-reproduction-Greece-sub}
    \end{subfigure}
\caption[Reproduction of regions and occupations in EU-SILC Greece]{Reproduction of regions and occupations at NUTS-1 level from the EU-SILC Greece. Thessaloniki municipality belongs to Region 3.}
  \label{fig:region-reproduction-Greece-main}
\end{figure}

\end{subappendices}